\documentclass{wmfki_arxiv}
\makeatletter
\jmlrproceedings{}{}
\jmlrvolume{}
\jmlrpages{}
\jmlryear{}
\renewcommand{\@titlefoot}{}
\makeatother
\usepackage{manuscript}

\begin{document}
\maketitle
\begin{abstract}
We propose mesh-free weak-form regression for learning interaction kernels from noisy particle trajectories. Our method uses smooth test functions to avoid differentiating noisy data and empirical particle averages to avoid density reconstruction and spatial grids. We then fit sparse combinations of candidate functions, evaluating interaction terms on sampled particle pairs to avoid all-pair quadratic cost. By transferring identifiability from the mean-field limit, we prove high-probability coefficient error bounds that separate finite-particle, time-discretization, observation-noise, and pair-sampling errors, without grid-resolution or density-estimation terms. We evaluate one-dimensional (1D) Newtonian-repulsion, 2D predator--prey, and 10D distance-localized experiments. The numerical results show trends consistent with the theoretical coefficient-error bounds. Comparisons of weak and forward-difference formulations with LASSO, Ridge, and OLS show that weak LASSO gives the smallest coefficient and drift errors in the three nominal experiments.
\end{abstract}

\begin{keywords}
Interacting particle systems, interaction-kernel learning, weak-form regression, sparse identification, particle-pair sampling.
\end{keywords}

\section{Introduction}
Interacting particle systems model collective behavior in biological and engineered systems, from animal groups and pedestrian crowds to robotic swarms and interacting cell populations \citep{Vicsek2012Collective,Mehes2014Collective}. Simple local interactions can generate macroscopic aggregation, flocking, milling, and synchronization, making their recovery important for interpretation, prediction, and control. We identify an unknown pairwise interaction kernel in a stochastic particle system from discrete position observations: unlike the forward problem, the trajectories are observed and the kernel is unknown.

One route infers interactions at the microscopic particle level. Available approaches include variational and nonparametric kernel estimators \citep{BonginiInferring2017,Lu2019Nonparametric,Lu2021Learninga}, likelihood-based inference for interacting diffusions \citep{Chen2021Maximum,Sharrock2023Online,Yao2022Meanfield}, and dictionary-based sparse regression for interpretable models \citep{Brunton2016Discovering,ManganInferring2016,CasadiegoModelfree2017}. These methods avoid a spatial representation of the population distribution but commonly use estimated velocities, drifts, or short-time increments, which can be sensitive to the observation interval and stochastic fluctuations \citep{Messenger2021Weaka,ReinboldUsing2020}. For a population of $N$ particles, evaluating candidate features over all ordered particle pairs costs $O(N^2)$ per observation time \citep{Jin2020Random}.

A second route uses the mean-field limit. Under standard Lipschitz and moment assumptions, the empirical distribution converges to the law of a McKean--Vlasov process and, when a density exists, to its nonlinear Fokker--Planck equation \citep{Sznitman1991Topics,Meleard1996Asymptotic,Chaintron2022Propagationa}. This connection motivates kernel inference from mean-field equations \citep{Lang2022Learning,CarrilloSparse2025a,shiExtractinginteractionkernels2025}. Weak-form regression transfers derivatives from data to smooth test functions by integration by parts, avoiding particle-level velocity or drift estimates \citep{Messenger2021Weaka,ReinboldUsing2020}. Using a density reconstructed from particle data, \citet{MessengerLearning2022a} evaluate the weak-form features by fast Fourier transform (FFT) convolutions on a spatial grid instead of particle pairs. This can be efficient for large populations and manageable grids, but density reconstruction and spatial discretization become costly in high dimension.

These two routes suggest combining their complementary strengths. We propose mesh-free weak-form regression directly from discrete particle trajectories: it retains the weak-form treatment of time derivatives while evaluating spatial terms from empirical particle averages. The resulting regression accommodates stochastic dynamics without numerical differentiation, density reconstruction, or a spatial grid. We sample particle pairs only after observing the trajectories, reducing feature-computation cost without altering the forward particle dynamics. The main contributions are:
\begin{itemize}
  \item We construct a weak-form estimator from discrete, possibly noisy particle positions. Applying It\^o's formula, empirical particle averaging, and time quadrature yields a regression that is linear in the dictionary coefficients, which we estimate by LASSO. Unlike an instantaneous drift fit, each regression row aggregates observations across particles and time, and every fitted quantity is computed from discrete positions. Sampling $P$ ordered pairs per observation time reduces pairwise computation from $O(N^2)$ to $O(P)$; the resulting design approximation is conditionally unbiased given the observed trajectories.
  \item We bound response and design perturbations from observation noise and pair sampling; with trapezoidal quadrature at observation interval $\Delta t$, the mean-square sampling contribution is $O(\Delta t/P)$. We relate the observed finite-particle design to a deterministic continuum design induced by the McKean--Vlasov marginal law, transfer singular-value and sparse-cone identifiability to finite designs, and obtain high-probability, non-asymptotic coefficient bounds for oracle least squares and LASSO. The bounds separate finite-particle process noise, time discretization, observation noise, and pair sampling, with no term for grid resolution or density estimation. They concern coefficient recovery rather than support recovery.
  \item We test the practical estimator in one-dimensional (1D) systems with a generic piecewise-linear dictionary that does not contain the true interaction and a sparse six-function dictionary containing the two functions used to generate the interaction and four redundant candidates. These experiments examine approximation, the scaling suggested by the bounds, observation noise, and pair sampling. We compare weak and forward-difference formulations with LASSO, Ridge, and OLS to assess the roles of the regression formulation and regularization. We also estimate three directed interactions in a two-dimensional (2D) predator--prey system and a sparse localized interaction in a ten-dimensional (10D) system.
\end{itemize}

\paragraph{Notation.}
Let $d$ denote the state dimension and $T>0$ the observation horizon.
We use $\E$ for expectation.
For an integer $r\ge0$, let $C_b^r(\R^d)$ be the space of functions in
$C^r(\R^d)$ with
$\|f\|_{C_b^r}:=\max_{0\le q\le r}\sup_{x\in\R^d}\|D^qf(x)\|<\infty$,
where derivative tensors carry their Euclidean operator norms. For an integer
$k\ge0$, the space $C_c^k((0,T))$ consists of $C^k$ functions compactly
supported in $(0,T)$.
We write $\nabla$ for the spatial gradient and $\Delta$ for the Laplacian, and use $\|\cdot\|$ for the Euclidean norm of vectors and the Frobenius norm of matrices unless another norm is indicated.

\section{Methodology}\label{sec:formulation}

Starting from a weak identity for the particle dynamics, we construct an
observed-data residual and use its squared norm as the regression loss.
We expand the kernel in a dictionary to obtain oracle ordinary least-squares
(OLS) and LASSO estimators, and sample particle pairs to reduce computational cost.
We evaluate every quantity in the practical estimator from the observed
particle positions.

\subsection{Particle Model and Inverse Problem}
\label{sec:particle-system}

We seek to recover an interaction kernel from discrete observations of $N$
particles governed on $[0,T]$, $T>0$, by the stochastic differential equations (SDEs)
\begin{equation}
  dX_i(t)=\frac{1}{N}\sum_{j\neq i}K^\star(X_i(t)-X_j(t))\,dt
  +\sigma\,dW_i(t),\qquad i=1,\dots,N.
  \label{eq:SDE}
\end{equation}
Here $X_i(t)\in\R^d$, $K^\star:\R^d\to\R^d$ denotes the
\emph{true} interaction kernel, $\sigma\ge0$ is the constant process-noise
intensity, and $\{W_i(t)\}$ are independent standard $d$-dimensional Brownian
motions. With square-integrable i.i.d. initial data and a globally Lipschitz $K^\star$ of at most linear growth (Assumption~\ref{ass:particle-system}), the system has a unique strong solution and uniform second-moment bounds (Lemma~\ref{lem:IPS_moments}).

Let $N_t\ge1$ be the number of observation intervals. The available data are particle positions observed at $t_n=n\Delta t$, $n=0,\ldots,N_t$, with $N_t\Delta t=T$, and possible observation errors:
\begin{equation}
  Y_i(t_n)
  =X_i(t_n)+\varepsilon^{\mathrm{obs}}_{i,n},
  \qquad i=1,\ldots,N,\quad n=0,\ldots,N_t,
  \label{eq:observation-model}
\end{equation}
Here $X_i(t_n)$ are the latent positions and
$\varepsilon^{\mathrm{obs}}_{i,n}$ are the observation errors. We write $Y_i^n:=Y_i(t_n)$ below.
The available data consist of the positions $\{Y_i(t_n)\}_{i,n}$ and their
sampling times. The latent positions, Brownian paths, and realized observation
errors are not observed. We assume that the process-noise intensity $\sigma$
is known, whereas the observation-noise scale $\sigma_{\mathrm{obs}}$ need not be
known. Exact position observations correspond to $\sigma_{\mathrm{obs}}=0$.

Unlike the forward problem, which predicts particle dynamics from a given
kernel, the inverse problem treats these observations as given and seeks
$K^\star$. We next construct a fitting criterion from the governing equations
without estimating particle velocities or reconstructing a population density.

\subsection{Weak-Form Regression}
\label{sec:regression-representation}

We start from the standard empirical-measure weak identity for the
particle dynamics \eqref{eq:SDE}, also used by
\citet[Lemma~2 and Appendix~A.4]{MessengerLearning2022a}. We state it below
for our spatial and temporal tests by applying a spatial test function,
averaging over the particles, and testing in time. We use this relation
to construct the fitting criterion.

\paragraph{Finite-particle weak identity.}
For trajectories solving~\eqref{eq:SDE} and fixed, trajectory-independent tests $\varphi\in C_b^2(\R^d)$ and $\psi\in C_c^1((0,T))$, It\^o's formula, particle averaging, and integration by parts give
\begin{align}
  \begin{aligned}
    0={}&\frac1N\sum_{i=1}^N\int_0^T \varphi(X_i(t))\,\dot\psi(t)\,dt
    +\frac1{N^2}\sum_{i\ne j}\int_0^T
    \nabla\varphi(X_i(t))\cdot K^\star(X_i(t)-X_j(t))\psi(t)\,dt \\
    &+\frac{\sigma^2}{2N}\sum_{i=1}^N\int_0^T
    \Delta\varphi(X_i(t))\psi(t)\,dt
    +\frac{\sigma}{N}\sum_{i=1}^N\int_0^T
    \psi(t)\nabla\varphi(X_i(t))\cdot dW_i(t).
  \end{aligned}
  \label{eq:particle-weak-exact}
\end{align}
Equation~\eqref{eq:particle-weak-exact} holds almost surely, and its stochastic integral has mean zero. We give the derivation in Appendix~\ref{sec:martingale-identity}. The first term in \eqref{eq:particle-weak-exact} measures the change in
the particle-averaged observable
$N^{-1}\sum_i\varphi(X_i(t))$ in weak form. The next two terms describe the effects of interactions and diffusion; the last
term accounts for stochastic fluctuations. The diffusion contribution is the
It\^o correction, whose coefficient is fixed by the known $\sigma$.
This relation motivates fitting a candidate kernel by the mismatch between
the observed change and the model terms.

\paragraph{Observed discrepancy.}
Choose spatial tests $\varphi_\ell\in C_b^2(\R^d)$,
$\ell=1,\ldots,L_x$, and temporal tests $\psi_k\in C_c^1((0,T))$,
$k=1,\ldots,L_t$. Index their $L:=L_xL_t$ combinations by $a=(\ell,k)$,
or equivalently by the stacked index $a=1,\ldots,L$.
Let $w_n$ be quadrature weights on the observation grid. For the composite
trapezoidal rule used in our implementation, $w_0=w_{N_t}=\Delta t/2$ and
$w_n=\Delta t$ for $1\le n\le N_t-1$.
For a candidate kernel $K$, replace the latent positions in the deterministic
terms of \eqref{eq:particle-weak-exact} by the observations $Y_i(t_n)$ and
apply time quadrature. The observed weak change minus the model terms defines
the residual vector $\eta_Y^{N,\Delta t}(K)\in\R^L$, whose entries are
\begin{equation}
  \begin{aligned}
    \eta_{Y,a}^{N,\Delta t}(K)
    :={}&-\frac1{\sqrt L\,N}\sum_{i=1}^N\sum_{n=0}^{N_t}w_n
    \left[\varphi_\ell(Y_i^n)\dot\psi_k(t_n)
    +\frac{\sigma^2}{2}\Delta\varphi_\ell(Y_i^n)\psi_k(t_n)\right] \\
    &-\frac1{\sqrt L\,N^2}\sum_{i\ne j}\sum_{n=0}^{N_t}
    w_n\psi_k(t_n)\nabla\varphi_\ell(Y_i^n)\cdot K(Y_i^n-Y_j^n).
  \end{aligned}
  \label{eq:weak-model-data-discrepancy}
\end{equation}
The factor $1/\sqrt L$ makes the squared norm an average over the test combinations. We define the risk by $\mathcal R_{N,\Delta t}(K;\{Y_i(t_n)\}_{i,n}):=\frac12\|\eta_Y^{N,\Delta t}(K)\|_2^2$. This criterion measures consistency with the weak evolution of the observed particle system. At the true kernel, the residual reflects stochastic dynamics, time discretization, and observation errors.

\paragraph{Dictionary representation.}
To obtain a finite-dimensional and interpretable kernel class, we fix vector-valued atoms $\Phi_m:\R^d\to\R^d$, $m=1,\ldots,M$, and write $K_\theta(z):=\sum_{m=1}^M\theta_m\Phi_m(z)$ for $\theta\in\R^M$. The dictionary expansion makes the interaction term in
\eqref{eq:weak-model-data-discrepancy} linear in $\theta$.
For a row $a=(\ell,k)$ and dictionary atom $m$, define the pairwise feature integrand $h_{a,m}(t,x,y):=\psi_k(t)\,\nabla\varphi_\ell(x)\cdot\Phi_m(x-y)$. Define the observed response vector $b_Y^{N,\Delta t}\in\R^L$ and design matrix $A_Y^{N,\Delta t}\in\R^{L\times M}$ by
  \begin{align}
    b_{Y,a}^{N,\Delta t}
    &:=-\frac1{\sqrt L\,N}\sum_{i=1}^N\sum_{n=0}^{N_t}w_n
    \left[\varphi_\ell(Y_i^n)\dot\psi_k(t_n)
    +\frac{\sigma^2}{2}\Delta\varphi_\ell(Y_i^n)\psi_k(t_n)\right],
    \label{eq:def-bY-dt} \\
    A_{Y,a,m}^{N,\Delta t}
     &:=\frac{1}{\sqrt{L}\,N^2}\sum_{i\ne j}\sum_{n=0}^{N_t}
    w_n\,h_{a,m}(t_n,Y_i^n,Y_j^n).
    \label{eq:def-AY-dt}
  \end{align}
By linearity, $\eta_Y^{N,\Delta t}(K_\theta)=b_Y^{N,\Delta t}-A_Y^{N,\Delta t}\theta$ for every $\theta\in\R^M$, and hence
\begin{equation}
  \mathcal R_{N,\Delta t}\bigl(K_\theta;\{Y_i(t_n)\}_{i,n}\bigr)
  =\frac12\|A_Y^{N,\Delta t}\theta-b_Y^{N,\Delta t}\|_2^2.
  \label{eq:weak-risk-regression}
\end{equation}
Equation~\eqref{eq:weak-risk-regression} expresses the observed risk as a least-squares loss whose matrix $A_Y^{N,\Delta t}$ averages over all ordered particle pairs. This identity holds for every $\theta$ and does not require the true kernel to belong to the dictionary span. We next reduce the cost of computing this matrix by sampling particle pairs before defining the estimators analyzed below.

\paragraph{Sampling particle pairs.}
\label{sec:random-pair}
Computing $A_Y^{N,\Delta t}$ from \eqref{eq:def-AY-dt} requires summing over
$N(N-1)$ ordered particle pairs, with $O(N^2)$ pairwise cost per observation
time. To reduce this cost to $O(P)$, we estimate these sums by sampling ordered pairs. Let $\mathcal F_Y=\sigma(Y_i(t_n):i=1,\ldots,N,\ n=0,\ldots,N_t)$. For $N\ge2$ and $P\in\N$, independently of $\mathcal F_Y$, we draw $P$ ordered off-diagonal index pairs uniformly with replacement at each $t_n$, independently across $n$, and denote the $\nu$th pair by $(I_{n,\nu},J_{n,\nu})$. The draws at $t_n$ are used for all tests and dictionary atoms. We set
\begin{equation}
  \bigl(\widetilde A_Y^{N,\Delta t,P}\bigr)_{a,m}
  :=\frac{N-1}{\sqrt L\,NP}\sum_{n=0}^{N_t}\sum_{\nu=1}^P
  w_n\,h_{a,m}(t_n,Y_{I_{n,\nu}}^n,Y_{J_{n,\nu}}^n).
  \label{eq:def-random-pair}
\end{equation}
Then $\E[\widetilde A_Y^{N,\Delta t,P}\mid\mathcal F_Y]=A_Y^{N,\Delta t}$; the factor $(N-1)/N$ matches the $N^{-2}$ normalization in $A_Y^{N,\Delta t}$. We write
$D_Y^{N,\Delta t,P}:=\widetilde A_Y^{N,\Delta t,P}-A_Y^{N,\Delta t}$
for the error in the matrix due to pair sampling. Pair sampling replaces $A_Y^{N,\Delta t}$ by $\widetilde A_Y^{N,\Delta t,P}$ and leaves $b_Y^{N,\Delta t}$ unchanged. Although $\widetilde A_Y^{N,\Delta t,P}$ is conditionally unbiased, squaring the residual introduces an additional contribution from pair sampling. The recovery theorem below controls its effect on coefficient estimation.

\paragraph{Defining the estimators.}
Replacing $A_Y^{N,\Delta t}$ by $\widetilde A_Y^{N,\Delta t,P}$ in the common squared loss, for $\lambda\ge0$ define
\begin{equation}
  \hat\theta_{Y,P}^\lambda
  \in\argmin_{\vartheta\in\R^M}
  \frac{1}{2}\bigl\|\widetilde A_Y^{N,\Delta t,P}\vartheta-b_Y^{N,\Delta t}\bigr\|_2^2
  +\lambda\|\vartheta\|_1.
  \label{eq:def-random-pair-LASSO}
\end{equation}
We call its minimizers weak LASSO for $\lambda>0$ and weak OLS for $\lambda=0$. These estimators are defined for any fixed dictionary and do not require $K^\star$ to belong to its span. Algorithm~\ref{alg:mesh-free-identification} in Appendix~\ref{app:oracle-complexity} summarizes the fitting workflow.

\paragraph{Computational cost.}

Sampling particle pairs makes construction of the regression system linear in $N$ when
$P=O(N)$, for a fixed observation grid, state dimension, dictionary, and test
family. Constructing $b_Y^{N,\Delta t}$ still requires averaging over all
particles and costs $O((N_t+1)N)$. Computing $A_Y^{N,\Delta t}$ evaluates $N(N-1)$
ordered pairs at each time, whereas computing $\widetilde A_Y^{N,\Delta t,P}$
evaluates only $P$. The corresponding costs are $O((N_t+1)N^2)$ and
$O((N_t+1)P)$. Thus the total cost of constructing the regression system from
sampled pairs is $O((N_t+1)(N+P))$. Sampling removes the quadratic pairwise work, but not
the linear work needed to form $b_Y^{N,\Delta t}$ from the observed particles. We describe solver costs and storage requirements in Appendix~\ref{app:oracle-complexity}.

\section{Main Recovery Result}
\label{sec:main-result}

We now establish high-probability coefficient-recovery bounds for weak LASSO and a theoretical oracle OLS estimator from noisy discrete particle observations with sampled pairs. We assume i.i.d. initial data satisfying $\E\|X_1(0)\|^4<\infty$, independent Brownian drivers, and a fixed dictionary with an exact, fixed-support representation $K^\star=K_{\theta^\star}$ of the globally Lipschitz, linearly growing true kernel. These conditions are not needed to define the estimators above. For a fixed $C_\varepsilon$, the observation errors satisfy $\E\|\varepsilon^{\mathrm{obs}}_{i,n}\|^2\le C_\varepsilon\sigma_{\mathrm{obs}}^2$ uniformly in $i$, $n$, $N$, and $\Delta t$; they are independent of the latent system but may be biased and mutually dependent. The fixed, trajectory-independent tests satisfy $\varphi_\ell\in C_b^3(\R^d)$ and $\psi_k\in C_c^1((0,T))$. The RMS quadrature error is $O((\Delta t)^p)$, with $\sum_{n=0}^{N_t}|w_n|\le C_W$ for a fixed $C_W$. See Assumptions~\ref{ass:particle-system}--\ref{ass:random-pair-sampling} in Appendix~\ref{sec:assumptions_A} for the formal conditions.

To define a deterministic reference for the finite-particle design, consider the McKean--Vlasov process associated with the mean-field limit of the particle system~\eqref{eq:SDE}, with $\rho_t:=\mathrm{Law}(\bar X(t))$ and $\bar X(0)$ distributed as $X_1(0)$. It solves
\begin{equation}
  d\bar X(t)
  =\left[\int_{\R^d}K^\star(\bar X(t)-y)\,\rho_t(dy)\right]dt
  +\sigma\,dW(t).
  \label{eq:mckean-vlasov-sde}
\end{equation}
Here $W$ is a standard $d$-dimensional Brownian motion independent of $\bar X(0)$. Under the conditions above, this nonlinear SDE has a unique strong solution and satisfies $\sup_{t\in[0,T]}\E\|\bar X(t)\|^4<\infty$; see Lemma~\ref{lem:MV-moment} in Appendix~\ref{app:mf-design}. Thus $\rho_t$ is a well-defined probability law, and the moment bound and dictionary/test regularity ensure finiteness of the deterministic continuum design
\begin{equation}
  A^\infty_{a,m}:=\frac1{\sqrt L}\int_0^T\int_{\R^d}\int_{\R^d}
  h_{a,m}(t,x,y)\,\rho_t(dx)\,\rho_t(dy)\,dt.
  \label{eq:main-continuum-design}
\end{equation}

The true-coefficient residual and the deviation of the sampled design from $A^\infty$ satisfy mean-square bounds with positive constants $C_\eta^Y$, $C_A^Y$, $C_{\mathrm{pair,Y}}^\eta$, and $C_{\mathrm{pair,Y}}^A$. These constants depend only on the fixed quantities in the assumptions and are uniform in $N$, $\Delta t$, and $P$; Propositions~\ref{prop:residual-design-estimates} and~\ref{prop:weak-design-convergence} establish the bounds. Set $Q_{\Delta t}:=\sum_{n=0}^{N_t}w_n^2$ and define the residual and design scales
\begin{align}
 \mathcal V^Y_{N,\Delta t,P}
 &=C_\eta^Y\!\left(\frac{\sigma^2}{N}+(\Delta t)^{2p}
 +\sigma_{\mathrm{obs}}^2\right)
 +C_{\mathrm{pair,Y}}^\eta\frac{Q_{\Delta t}}P,
 \label{eq:main-residual-scale}\\
 \mathcal D^Y_{N,\Delta t,P}
 &=C_A^Y\!\left(\frac1N+(\Delta t)^{2p}
 +\sigma_{\mathrm{obs}}^2\right)
 +C_{\mathrm{pair,Y}}^A\frac{Q_{\Delta t}}P.
 \label{eq:main-design-scale}
\end{align}
The first controls the true-coefficient residual; the second controls the Frobenius distance from the continuum design. Independence of the pair draws across observation times produces the factor $Q_{\Delta t}/P$. For trapezoidal weights, $Q_{\Delta t}\le T\Delta t$, so this contribution is $O(\Delta t/P)$. Thus, for fixed $P$, reducing $\Delta t$ decreases both $(\Delta t)^{2p}$ and the pair-sampling contribution.

To state the theorem, let $S:=\supp(\theta^\star)$, $s:=|S|$, and $S^c:=\{1,\ldots,M\}\setminus S$. For a matrix $B$, let $B_S$ denote the columns indexed by $S$; for a vector $v$, let $v_S$ denote its entries in $S$. To isolate coefficient estimation from support selection, define the theoretical oracle OLS estimator, which assumes that $S$ is known:
\begin{equation}
  \hat\theta_{S,Y,P}
  \in\argmin_{\vartheta\in\R^s}
  \frac12\bigl\|\widetilde A_{Y,S}^{N,\Delta t,P}\vartheta-b_Y^{N,\Delta t}\bigr\|_2^2.
  \label{eq:def-random-pair-oracle}
\end{equation}
For weak LASSO, define the cone $\mathcal C_S:=\{u:\|u_{S^c}\|_1\le3\|u_S\|_1\}$ and set $R_\infty:=\max_m\|A^\infty_{:,m}\|_2$, where $A^\infty_{:,m}$ is the $m$th column of $A^\infty$. For oracle OLS, write $\sigma_{\min}(B):=\inf_{\|v\|_2=1}\|Bv\|_2$. The theorem below analyzes oracle OLS and weak LASSO.

\begin{theorem}[Coefficient recovery with sampled particle pairs]
\label{thm:random-pair-recovery}
Under Assumptions~\ref{ass:particle-system}--\ref{ass:random-pair-sampling}, suppose additionally that the spatial tests have bounded derivatives through order three and $\E\|X_1(0)\|^4<\infty$. Let $\delta_A,\varepsilon\in(0,1)$ satisfy $\delta_A+\varepsilon<1$. Each conclusion below holds with probability at least $1-\delta_A-\varepsilon$.
\begin{enumerate}[label=(\roman*),leftmargin=2.1em,itemsep=.4em,topsep=.4em]
 \item \textbf{Oracle OLS (known support).} If there exists $\gamma_\infty>0$ such that $\sigma_{\min}(A_S^\infty)\ge\gamma_\infty$, and $\mathcal D^Y_{N,\Delta t,P}\le\delta_A\gamma_\infty^2/4$, then
 \begin{equation}
   \|\hat\theta_{S,Y,P}-\theta_S^\star\|_2^2
   \le\frac{4\mathcal V^Y_{N,\Delta t,P}}
   {\varepsilon\gamma_\infty^2}.
   \label{eq:main-oracle-recovery-bound}
 \end{equation}
 \item \textbf{LASSO (unknown support).} If there exists $\gamma_{\infty,\mathrm{sp}}>0$ such that
 $\inf_{u\in\mathcal C_S\setminus\{0\}}\|A^\infty u\|_2/\|u\|_2\ge\gamma_{\infty,\mathrm{sp}}$, and $\mathcal D^Y_{N,\Delta t,P}\le\delta_A\gamma_{\infty,\mathrm{sp}}^2/4$, choose $\lambda>0$ satisfying
 \begin{equation}
   \lambda\ge2\left(R_\infty+\frac{\gamma_{\infty,\mathrm{sp}}}{2}\right)
   \sqrt{\frac{\mathcal V^Y_{N,\Delta t,P}}{\varepsilon}}.
   \label{eq:random-pair-main-penalty}
 \end{equation}
 For every minimizer of~\eqref{eq:def-random-pair-LASSO}, set $u:=\hat\theta_{Y,P}^\lambda-\theta^\star$. Then
 \begin{equation}
   \|u_{S^c}\|_1\le3\|u_S\|_1,
   \qquad
   \|u\|_2^2\le\frac{144s\lambda^2}{\gamma_{\infty,\mathrm{sp}}^4}.
   \label{eq:main-lasso-recovery-bounds}
 \end{equation}
\end{enumerate}
\end{theorem}

Theorem~\ref{thm:random-pair-recovery} gives high-probability bounds on squared coefficient error from noisy position observations. The cone inequality controls error on inactive atoms relative to error on the true support. For fixed $\sigma$ and confidence levels, $1/N+(\Delta t)^{2p}+\sigma_{\mathrm{obs}}^2+Q_{\Delta t}/P\to0$ makes both $\mathcal D^Y_{N,\Delta t,P}$ and $\mathcal V^Y_{N,\Delta t,P}$ vanish. If the corresponding continuum identifiability condition holds with $\gamma_\infty>0$ or $\gamma_{\infty,\mathrm{sp}}>0$, then $\mathcal D^Y_{N,\Delta t,P}\to0$ eventually implies the associated finite-sample design requirement in Theorem~\ref{thm:random-pair-recovery}. The oracle OLS bound then tends to zero, and choosing equality in~\eqref{eq:random-pair-main-penalty} ensures the same for the LASSO bound. In the experiments, $\lambda$ is instead selected using an independent validation trajectory. \WMFKILeanStatement

\section{Numerical Experiments}
\label{sec:numerical-experiments}

We study the 1D Newtonian-repulsion, 2D predator--prey, and 10D distance-localized experiments. Weak regression uses the time-integrated test equations above, whereas strong forward-difference (strong FD) regression fits $[Y_i(t_{n+1})-Y_i(t_n)]/\Delta t$ to particle-drift features without temporal tests. For each formulation, OLS is unregularized, Ridge uses an $\ell_2$ penalty, and LASSO uses an $\ell_1$ penalty. We select the weak LASSO parameter $\lambda$ in~\eqref{eq:def-random-pair-LASSO} and the strong FD--LASSO and Ridge penalty parameters by minimizing the residual loss for the corresponding formulation on independent validation data. Appendix~\ref{app:additional-numerics} gives the candidate grids and complete protocols.

Each weak regression uses $L_x=25$ scalar Gaussian spatial test functions $\{\varphi_\ell:\R^d\to\R\}_{\ell=1}^{L_x}$ and $L_t=8$ smooth temporal test functions $\{\psi_k:(0,T)\to\R\}_{k=1}^{L_t}$; $\nabla\varphi_\ell$, $\Delta\varphi_\ell$, and $\dot\psi_k$ are evaluated in closed form, and the time integrals in~\eqref{eq:particle-weak-exact} are approximated by composite trapezoidal quadrature. The 1D and predator--prey tests are fixed before observation; 10D centers are data dependent. The timing measurements were obtained on a MacBook Pro with an Apple M4 Max chip and 128~GB of memory. \WMFKINumericalCodeStatement

\subsection{One-Dimensional Approximation with a Generic Piecewise-Linear Dictionary}
\label{sec:num-hat-approximation}

Following the one-dimensional Newtonian-repulsion model of \citet{MessengerLearning2022a}, we use $K^\star(z)=z\,\phi(|z|)$, where $\phi(r)=(r+\varepsilon_K)^{-1}-1$ and $\varepsilon_K=10^{-8}$.
We approximate the radial profile $\phi$ with a generic piecewise-linear dictionary on $[0,4]$. For dictionary size $M$, let $0=r_1<\cdots<r_M=4$ be uniformly spaced knots. The $m$th nodal basis function $b_m$ is the continuous piecewise-linear function that equals one at $r_m$ and zero at every other knot, so the coefficients are the fitted radial-profile values at the knots. We fit $K_\theta(z)=z\sum_{m=1}^M\theta_m b_m(|z|)$ and extend the radial profile constantly for $|z|>4$. The dictionary is not tailored to $K^\star$: because $\phi$ is not piecewise linear, the true interaction lies outside every finite span considered here.

We evaluate ten dictionary sizes $M$ over 100 repetitions with common seeds across sizes. Each repetition has independent training, validation, and latent test trajectories with $N=1000$, $T=2$, $\Delta t=0.005$, $\sigma=0.1$, $\sigma_{\mathrm{obs}}=0.01$, and $P=50N$. Independent observation noise is added to training and validation, and each LASSO penalty minimizes the corresponding validation loss.

\begin{figure}[t]
  \centering
  \includegraphics[width=\linewidth]{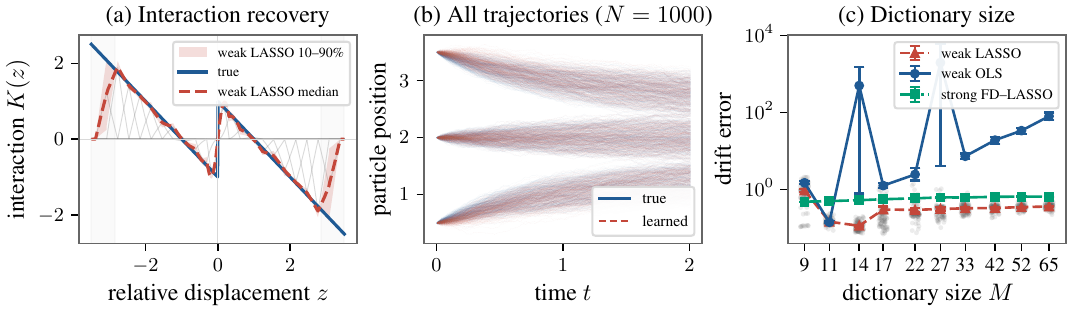}
  \caption{Results with the generic piecewise-linear dictionary. (a) True interaction and median weak LASSO fit at $M=14$, with a 10--90\% band over 100 repetitions; gray curves show median atom contributions and shading marks data-sparse tails. (b) True and weak LASSO trajectories generated from the same initial condition and Brownian increments. (c) Mean drift error for the three methods, with 95\% confidence intervals; gray points show the 100 weak LASSO errors.}
  \label{fig:one-dimensional-hat-practical}
\end{figure}

Figure~\ref{fig:one-dimensional-hat-practical}(a) compares the true interaction with the median weak LASSO fit and its 10--90\% band. The median fit captures the attraction--repulsion shape, with discrepancies in data-sparse tails. Figure~\ref{fig:one-dimensional-hat-practical}(b) displays the true and weak LASSO particle trajectories generated from the same initial condition and Brownian increments; their relative trajectory error is $0.0150$. Figure~\ref{fig:one-dimensional-hat-practical}(c) reports mean drift error against dictionary size for the three methods. Weak LASSO reaches its minimum, $0.113$, at $M=14$ and is best at eight of ten sizes. This is consistent with an approximation--estimation trade-off \citep{Lu2021Learninga,Pavliotis2026Fourier}: fixed noisy trajectories may not constrain added basis functions well. Every weak OLS matrix is rank deficient, and its errors become unstable; strong FD--LASSO remains finite but is better than weak LASSO only at $M=9$.

\FloatBarrier

\subsection{One-Dimensional Newtonian-Repulsion Experiment}
\label{sec:num-newtonian-repulsion-experiment}

To compare the numerical error trends with the separate error contributions in the recovery bounds, we use the same 1D particle model and true interaction $K^\star$, but replace the generic piecewise-linear dictionary with a dictionary that represents $K^\star$ exactly. Its scalar profiles are
$\{(r+\varepsilon_K)^{-1},1,e^{-r/0.25},e^{-r/0.5},e^{-r},e^{-r/2}\}$, with $\theta^\star=(1,-1,0,0,0,0)^\top$. Adapted from \citet{MessengerLearning2022a}, the first two profiles generate the true interaction and the other four are correlated inactive atoms.

\begin{figure}[t]
  \centering
  \includegraphics[width=\linewidth]{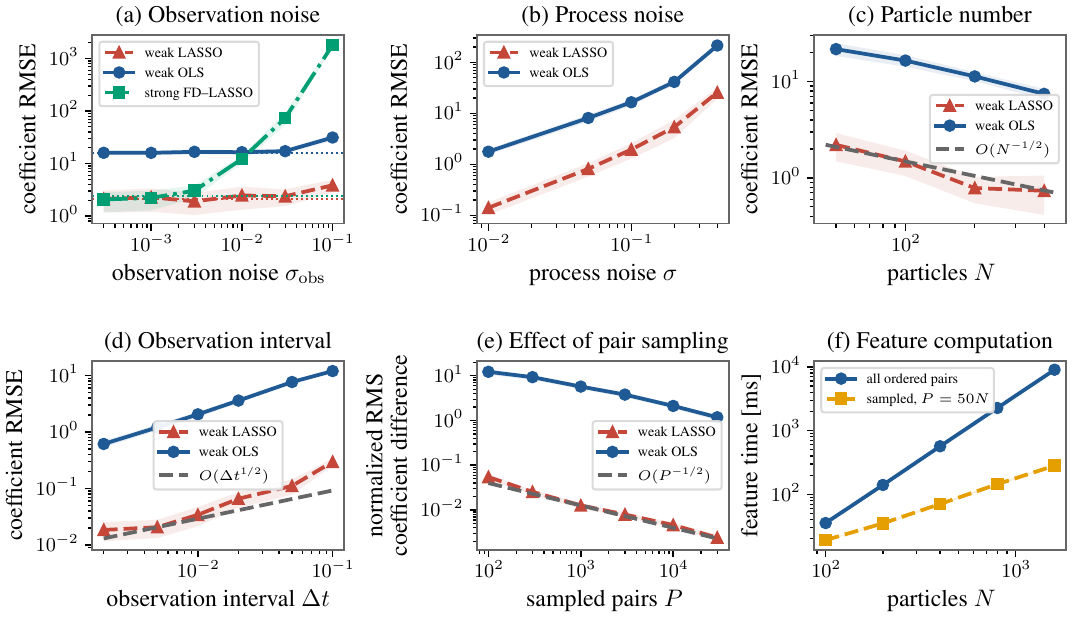}
  \caption{1D Newtonian-repulsion experiment. Red/blue curves are weak LASSO and weak OLS; panel~(a) also shows strong FD--LASSO in green. Panels~(a--d) use $P=50N$ and vary observation noise, process noise, particle number, and observation interval. Panel~(e) varies $P$; panel~(f) compares all-pair and sampled-feature times. Panels~(b--e) use exact positions. Bands show 95\% intervals in (a--e) and interquartile ranges in (f); gray lines give reference rates.}
  \label{fig:validation-sweeps}
\end{figure}

Figure~\ref{fig:validation-sweeps}(a) reports coefficient RMSE versus the observation-noise level $\sigma_{\mathrm{obs}}$ in~\eqref{eq:observation-model} for weak LASSO, weak OLS, and strong FD--LASSO. Their RMSEs are $2.09$, $15.99$, and $2.42$ with exact positions, and $3.85$, $31.28$, and $1814$ at $\sigma_{\mathrm{obs}}=10^{-1}$. Time integration in the weak formulation limits noise amplification, while sparse regularization reduces error from redundant columns.

Figure~\ref{fig:validation-sweeps}(b) reports coefficient RMSE versus the process-noise level $\sigma$, whereas Figure~\ref{fig:validation-sweeps}(c) reports it versus particle number $N$; both use exact positions. Weak LASSO has a smaller coefficient RMSE than weak OLS at every setting. For both estimators, the error is approximately linear in $\sigma$ at lower values and increases more rapidly at the largest values. For weak LASSO, the lower-$\sigma$ behavior agrees with the coefficient-error scale obtained from the $\sigma^2/N$ term in~\eqref{eq:main-residual-scale} through~\eqref{eq:random-pair-main-penalty}--\eqref{eq:main-lasso-recovery-bounds} when that term dominates. At fixed $\sigma$, the same scale is $N^{-1/2}$; the fitted weak LASSO slope in Figure~\ref{fig:validation-sweeps}(c) is $-0.567$. Figure~\ref{fig:validation-sweeps}(d) reports coefficient RMSE versus the observation interval with $\sigma=\sigma_{\mathrm{obs}}=0$, so fixed noise terms are absent. For weak LASSO and $\Delta t\le0.02$, the fitted slope is $0.555$. This agrees with the $O(\!\sqrt{\Delta t/P})$ scale obtained from the $O(\Delta t/P)$ mean-square pair-sampling term in~\eqref{eq:main-residual-scale}.

Figure~\ref{fig:validation-sweeps}(e) reports the normalized RMS coefficient difference from each estimator's all-pair fit versus $P$. For each repetition, sampled validation at $P_0=10^4$ selects one absolute LASSO penalty, which is fixed across all $P$ and also used for the all-pair reference. The fitted slopes are $-0.528$ for weak LASSO and $-0.413$ for weak OLS. The weak LASSO slope is close to the $P^{-1/2}$ scale implied by the mean-square terms in~\eqref{eq:main-residual-scale}--\eqref{eq:main-design-scale}. Figure~\ref{fig:validation-sweeps}(f) reports all-pair and $P=50N$ sampled feature-construction times versus $N$. All-pair assembly is approximately quadratic in $N$; sampled assembly is nearly linear and $32$ times faster at $N=1600$.

\FloatBarrier

\subsection{Two- and Ten-Dimensional Applications}
\label{sec:numerical-summary}

The 2D predator--prey experiment adapts the predator--swarm model of \citet{ChenKolokolnikov2014Predator}, using short-range prey repulsion, longer-range prey attraction, and directed cross-species interactions. Across 20 repetitions at $\sigma_{\mathrm{obs}}=0.01$, the median drift errors are $0.0603$ for weak LASSO and $0.174$ for weak OLS. The learned weak LASSO system reproduces the ring pattern over $[0,3]$ with trajectory error $0.148$. Drift error increases with observation noise and the observation interval, while the single predator limits gains from adding prey. In the nominal comparison, weak LASSO has the smallest errors.

The 10D distance-localized experiment uses a sparse radial interaction with localized attractive and repulsive features. Across 20 repetitions at $\sigma_{\mathrm{obs}}=0.01$ and $P=50N$, the mean weak LASSO coefficient error grows with process noise and decreases with particle and pair counts, with log--log slopes $1.294$, $-0.873$, and $-0.370$; it remains below weak OLS throughout. All weak methods have smaller nominal errors than their strong FD counterparts.

At the common nominal level $\sigma_{\mathrm{obs}}=0.01$, Table~\ref{tab:cross-example-comparison} compares weak and strong FD formulations with LASSO, Ridge, and OLS. Weak LASSO has the smallest coefficient and drift RMSE in all three systems; the 1D row uses the realizable six-atom dictionary.

\begin{table}[!ht]
  \centering
  \caption{Coefficient and drift RMSEs; penalties are selected on independent validation data.}
  \label{tab:cross-example-comparison}
  \footnotesize
  \setlength{\tabcolsep}{1.5pt}
  \begin{tabular*}{\linewidth}{@{\extracolsep{\fill}}lcccccc@{}}
    \toprule
    & \multicolumn{3}{c}{Weak form} & \multicolumn{3}{c}{Strong FD} \\
    \cmidrule(lr){2-4}\cmidrule(lr){5-7}
    System & LASSO & Ridge & OLS & LASSO & Ridge & OLS \\
    \midrule
    1D Newtonian & \textbf{2.47/0.0444} & 5.58/0.0594 & 16.5/0.112 & 12.2/0.501 & 13.6/0.502 & 26.6/0.518 \\
    2D predator--prey & \textbf{0.347/0.224} & 134/0.277 & 193/0.363 & 99.1/1.15 & 1740/1.19 & 1810/1.53 \\
    10D localized & \textbf{0.155/0.0984} & 0.305/0.142 & 0.277/0.123 & 0.569/0.305 & 0.558/0.301 & 0.601/0.314 \\
    \bottomrule
  \end{tabular*}
\end{table}

\section{Conclusion}
\label{sec:conclusion}

We introduced mesh-free weak-form regression that learns interaction kernels from particle positions without estimating velocities, densities, or spatial grids. Sampling $P$ pairs reduces feature construction from $O(N^2)$ to $O(P)$. The coefficient bounds separate the effects of particle number, time quadrature, observation noise, and pair sampling. The 1D Newtonian-repulsion experiments covered generic approximation and realizable recovery with scaling consistent with these bounds. The 2D predator--prey experiment recovered three directed interactions and ring dynamics, while the 10D distance-localized experiment recovered a sparse interaction near two pairwise distances. Across these comparisons, weak LASSO had the smallest coefficient and drift errors in all three nominal experiments. Support recovery and theory for heterogeneous systems remain open.

\subsection*{Generative AI Use}
Generative AI tools were used to assist with mathematical proofs, numerical computations, and writing. The authors reviewed and verified all AI-assisted content and take responsibility for the final content of the paper.
\clearpage
\acks{D. Kalise is supported by the Air Force Office of Scientific Research under award number FA8655-26-1-B011.}

\appendix
\section{Practical Algorithm and Computational Details}
\label{app:oracle-complexity}

We summarize the practical finite-particle workflow in
Algorithm~\ref{alg:mesh-free-identification}. We use LASSO in the practical
procedure and retain the oracle OLS estimator in~\eqref{eq:def-random-pair-oracle}
as a theoretical reference.

\RestyleAlgo{ruled}
\begin{algorithm2e}
  \caption{Mesh-free weak-form interaction-kernel identification}
  \label{alg:mesh-free-identification}
  \begin{algorithmic}[1]
    \REQUIRE Observations $\{Y_i(t_n)\}$; known $\sigma$; atoms
      $\{\Phi_m\}_{m=1}^M$; spatial and temporal tests; quadrature weights;
      pair count $P$; penalty $\lambda$.
    \STATE Evaluate $b_Y^{N,\Delta t}$ from empirical particle averages using
      \eqref{eq:def-bY-dt}.
    \STATE At each $t_n$, draw $P$ ordered particle pairs with replacement,
      independently across $n$ and independently of the observations.
    \STATE Use each time's draws for all tests and atoms to assemble
      $\widetilde A_Y^{N,\Delta t,P}$ using \eqref{eq:def-random-pair}.
    \STATE Solve the LASSO problem \eqref{eq:def-random-pair-LASSO} for
      $\hat\theta_{Y,P}^\lambda$.
    \ENSURE Estimated coefficients $\hat\theta_{Y,P}^\lambda$ and fitted kernel
      $K_{\hat\theta_{Y,P}^\lambda}$.
  \end{algorithmic}
\end{algorithm2e}

\paragraph{Solver cost.}
The optimization problem has fixed dimensions as $N$ and $N_t$ increase,
provided the dictionary and test family remain fixed. We sum over particles
and observation times before solving the regression, so $b_Y^{N,\Delta t}$ has
$L$ entries and $\widetilde A_Y^{N,\Delta t,P}$ has $L\times M$ entries.
For LASSO, cyclic coordinate descent costs $O(LM)$ per sweep: each of the
$M$ coordinate updates uses a column of length $L$ and updates the residual.
The number of sweeps depends on conditioning and the stopping tolerance.
The linear cost of computing the regression system therefore does not by itself give a uniform linear
bound on the total runtime.

\paragraph{Storage.}
Computing the pairwise features also need not require quadratic storage in $N$. We can
accumulate the sums defining $b_Y^{N,\Delta t}$ and $\widetilde A_Y^{N,\Delta t,P}$ without storing a
feature array for all particle pairs. The completed regression system requires
$O(LM+L)$ storage, whether $A_Y^{N,\Delta t}$ is computed from all ordered pairs
or $\widetilde A_Y^{N,\Delta t,P}$ from $P$ sampled pairs, excluding
temporary workspace and the observed trajectories. Keeping all observations
requires $O((N_t+1)Nd)$ storage. The computational saving from pair sampling
is therefore a reduction in feature evaluations, not in the dimensions of the
completed regression system.

\section{Detailed Assumptions}
\label{app:theory-details}
\label{sec:assumptions_A}

We consider the interacting particle system~\eqref{eq:SDE} on
$[0,T]$ and the observed regression in Section~\ref{sec:regression-representation}.
Throughout the analysis, the state dimension $d$ is fixed. Fix a finite
upper bound $\overline\sigma_{\mathrm{obs}}\ge0$ for the observation-noise scale.
Unless stated
otherwise, constants may depend on $d$, the fixed model and dictionary, the test
functions, the time horizon, fixed moment bounds, and
$\overline\sigma_{\mathrm{obs}}$. All such constants are uniform in $N$,
$\Delta t$, $P$, and
$\sigma_{\mathrm{obs}}\in[0,\overline\sigma_{\mathrm{obs}}]$.

\begin{assumption}[Particle system]
  \label{ass:particle-system}
  \begin{enumerate}[label=(\roman*),ref=\thetheorem(\roman*),leftmargin=2.2em]
    \item \label{ass:particle-initial}
          \textbf{Initial data.} The initial conditions
          $\{X_i(0)\}_{i=1}^N$ are i.i.d.\ with common law $\rho_0$ and satisfy
          \[
            \E\|X_1(0)\|^2<\infty.
          \]
    \item \label{ass:particle-noise}
          \textbf{Driving noises.} The driving noises $\{W_i\}_{i=1}^N$ are
          independent standard $d$-dimensional Brownian motions, independent
          of the initial conditions.
    \item \label{ass:particle-kernel}
          \textbf{Interaction kernel.} The true interaction kernel
          $K^\star:\R^d\to\R^d$ is globally Lipschitz and of at most linear
          growth: there exist
          $L_K,G_K\ge0$ such that for all $x,y\in\R^d$,
          \[
            \|K^\star(x)-K^\star(y)\|\le L_K\|x-y\|,
            \qquad \|K^\star(x)\|\le G_K(1+\|x\|).
          \]
  \end{enumerate}
\end{assumption}

\begin{assumption}[Observation errors]
  \label{ass:observation-noise}
  The family of observation errors
  $\{\varepsilon^{\mathrm{obs}}_{i,n}:1\le i\le N,\ 0\le n\le N_t\}$
  is independent of the particle initial data and driving Brownian motions.
  For a fixed constant $C_\varepsilon>0$, the errors take values in $\R^d$ and satisfy
  \[
    \E\bigl\|\varepsilon^{\mathrm{obs}}_{i,n}\bigr\|^2
    \le C_\varepsilon\sigma_{\mathrm{obs}}^2,
    \qquad
    0\le\sigma_{\mathrm{obs}}\le\overline\sigma_{\mathrm{obs}},
  \]
  uniformly in $i,n,N$, and $\Delta t$.
\end{assumption}

\begin{assumption}[Dictionary model]
  \label{ass:dictionary-model}
  \begin{enumerate}[label=(\roman*),ref=\thetheorem(\roman*),leftmargin=2.2em]
    \item \label{ass:dictionary-regularity}
          \textbf{Regularity.} The dictionary atoms $\{\Phi_m\}_{m=1}^M$ map
          $\R^d$ to $\R^d$ and are globally Lipschitz and of at most linear
          growth uniformly in
          $m$: there exist $L_\Phi,G_\Phi\ge0$ such that for all $m$ and all
          $x,y\in\R^d$,
          \[
            \|\Phi_m(x)-\Phi_m(y)\|\le L_\Phi\|x-y\|,
            \qquad \|\Phi_m(x)\|\le G_\Phi(1+\|x\|).
          \]
          The dictionary size $M$ is fixed and does not depend on $N$,
          $\Delta t$, or $P$.
    \item \label{ass:dictionary-realizability}
          \textbf{Realizability and fixed support.} The exact representation
          stated in Section~\ref{sec:main-result} holds. Recall the active set
          $S=\supp(\theta^\star)$ and its size $s=|S|$. The
          coefficient vector $\theta^\star$, its support $S$, and $s$ are
          fixed and do not depend on $N$, $\Delta t$, or $P$.
  \end{enumerate}
\end{assumption}

\begin{assumption}[Fixed admissible test functions]
  \label{ass:test_functions_A}
  Fix an integer $r\ge 2$, and fix the numbers $L_x$ and $L_t$ of
  spatial and temporal test functions, with $L=L_xL_t$.
  The spatial and temporal test functions are deterministic and fixed in advance:
  they do not depend on the observed trajectories, on $N$, or on $\Delta t$.
  For each spatial index $\ell=1,\dots,L_x$ and temporal index $k=1,\dots,L_t$,
  the test functions $\varphi_\ell$ and $\psi_k$ satisfy:
  \begin{enumerate}[label=(\roman*),leftmargin=2.2em]
    \item \textbf{Spatial tests:}
          $\varphi_\ell\in C_b^{r}(\R^d)$, and there exists $C_{\varphi}>0$
          such that
          \[
            \sup_{1\le\ell\le L_x}\|\varphi_\ell\|_{C_b^r}
            \le C_{\varphi}.
          \]
    \item \textbf{Temporal tests:}
          \[
            \psi_k\in C_c^1((0,T)).
          \]
          Moreover, there exists $C_{\psi}>0$ such that uniformly in $k$,
          \[
            \|\psi_k\|_{L^\infty(0,T)}
            +\|\dot\psi_k\|_{L^\infty(0,T)}
            \le C_{\psi}.
          \]
  \end{enumerate}
\end{assumption}

To state the quadrature assumption, define the latent discrete response
vector $b^{N,\Delta t}\in\R^L$ and design matrix
$A^{N,\Delta t}\in\R^{L\times M}$ by
\begin{equation}
  b^{N,\Delta t}_a
  :=-\frac1{\sqrt L\,N}\sum_{i=1}^N\sum_{n=0}^{N_t}w_n
  \left[\varphi_\ell(X_i(t_n))\,\dot\psi_k(t_n)
  +\frac{\sigma^2}{2}\Delta\varphi_\ell(X_i(t_n))\,\psi_k(t_n)\right].
  \label{eq:def-b-dt}
\end{equation}
\begin{equation}
  A^{N,\Delta t}_{a,m}
  :=\frac1{\sqrt L\,N^2}\sum_{i\ne j}\sum_{n=0}^{N_t}
    w_n\,h_{a,m}(t_n,X_i(t_n),X_j(t_n)).
  \label{eq:def-A-dt}
\end{equation}
These use the same tests, dictionary, and quadrature weights as the observed
quantities \eqref{eq:def-bY-dt}--\eqref{eq:def-AY-dt}.
Their continuous-time counterparts $b^{N,\mathrm{cts}}\in\R^L$ and
$A^{N,\mathrm{cts}}\in\R^{L\times M}$ are defined entrywise by
\begin{equation}
  b^{N,\mathrm{cts}}_a
  :=-\frac1{\sqrt L\,N}\sum_{i=1}^N\int_0^T
  \left[\varphi_\ell(X_i(t))\,\dot\psi_k(t)
  +\frac{\sigma^2}{2}\Delta\varphi_\ell(X_i(t))\,\psi_k(t)\right]dt,
  \label{eq:def-b-cts}
\end{equation}
\begin{equation}
  A^{N,\mathrm{cts}}_{a,m}
  :=\frac1{\sqrt L\,N^2}\sum_{i\ne j}\int_0^T
    h_{a,m}(t,X_i(t),X_j(t))\,dt.
  \label{eq:def-A-cts}
\end{equation}
Here $a=(\ell,k)$ and $h_{a,m}$ is the pairwise integrand defined in Section~\ref{sec:regression-representation}.

\begin{assumption}[Time discretization and quadrature accuracy]
  \label{ass:time_grid_A}
  For the discrete approximation \eqref{eq:def-b-dt}--\eqref{eq:def-A-dt},
  there exist constants $p>0$ and $C_{\mathrm{quad}}>0$, independent of $N$,
  $\Delta t$, and $L$ for the fixed test family and fixed dictionary, such that
  \[
    \E\left[\bigl\|b^{N,\Delta t}-b^{N,\mathrm{cts}}\bigr\|_2^2\right]
    \le C_{\mathrm{quad}}\,(\Delta t)^{2p},
    \quad
    \E\left[\bigl\|A^{N,\Delta t}-A^{N,\mathrm{cts}}\bigr\|_F^2\right]
    \le C_{\mathrm{quad}}\,(\Delta t)^{2p}.
  \]
  In addition, the quadrature weights satisfy
  \[
    W_{\Delta t}:=\sum_{n=0}^{N_t}|w_n|\le C_W
  \]
  for a constant $C_W$ independent of $N$ and $\Delta t$.
\end{assumption}

The exponent $p$ in Assumption~\ref{ass:time_grid_A} describes the
root-mean-square accuracy of the time-discrete approximations for the
stochastic trajectory integrands. It is independent of the $C^1$ regularity
imposed on the temporal test functions and does not follow solely from the
deterministic order of the quadrature rule.
For sufficiently regular It\^o-semimartingale integrands, one may expect
$p=1$ for composite trapezoidal quadrature, whereas higher rates may be
possible for smooth deterministic integrands. Establishing a quadrature rate
for all integrands appearing in the present regression would require
additional regularity assumptions on the spatial test functions and
dictionary atoms and is left for future work. For the composite trapezoidal
rule, $W_{\Delta t}=T$, so the total-weight condition is automatic.

Computing $\widetilde A_Y^{N,\Delta t,P}$ introduces an auxiliary source of randomness in
addition to the initial data and Brownian motions of the particle system. The
following assumption formalizes this sampling mechanism and its independence
from the observed trajectories.

\begin{assumption}[Pair sampling]
  \label{ass:random-pair-sampling}
  Assume $N\ge2$ and $P\in\N$. The index pairs $(I_{n,\nu},J_{n,\nu})$ in
  Section~\ref{sec:random-pair} are sampled with replacement, independently
  across $(n,\nu)$, and independently of
  $\mathcal F_Y$.
\end{assumption}

For the composite trapezoidal weights from
Section~\ref{sec:regression-representation},
\[
  Q_{\Delta t}
  =2\left(\frac{\Delta t}{2}\right)^2
  +(N_t-1)(\Delta t)^2
  =T\Delta t-\frac{(\Delta t)^2}{2}.
\]
Independent redraws across observation times remove cross-time covariance
terms from the conditional variance. Reusing the same pairs across times
would instead generally leave an $O(P^{-1})$ contribution.

\section{Proof of Theorem~\ref{thm:random-pair-recovery} and Supporting Results}
\label{app:proofs}

In the appendices, $\mathbb{P}$ denotes probability and $\mathbf 1_E$ the indicator of an event $E$. For a finite measure $\mu$ on $\R^d$ and a $\mu$-integrable function $f$, we write
$\ip{\mu}{f}:=\int_{\R^d}f(x)\,\mu(dx)$.

\subsection{Finite-Particle Weak Identity}
\label{sec:martingale-identity}
\begin{proof}[Derivation of \eqref{eq:particle-weak-exact}]

The empirical measure of the particle system and its pairing with a spatial test
function $\varphi$ are
\[
  \mu_t^N := \frac{1}{N}\sum_{i=1}^N \delta_{X_i(t)},
  \qquad
  \ip{\mu_t^N}{\varphi} = \frac{1}{N}\sum_{i=1}^N \varphi(X_i(t)).
\]
Applying It\^o's formula \cite[Theorem~3.3.6]{Karatzas1998Brownian} to $\varphi(X_i(t))$ along
\eqref{eq:SDE} for a spatial test function $\varphi\in C_b^2(\R^d)$ and averaging over $i$ yields the
semimartingale identity
\begin{equation}
  d\ip{\mu_t^N}{\varphi}
  =\frac{1}{N^2}\sum_{i\neq j}
  \nabla\varphi(X_i(t))\cdot K^\star(X_i(t)-X_j(t))\,dt
  +\frac{\sigma^2}{2}\,\ip{\mu_t^N}{\Delta\varphi}\,dt
  +dM_t^{N,\varphi},
  \label{eq:empirical-martingale}
\end{equation}
where
\begin{equation}
  M_t^{N,\varphi}
  :=
  \frac{\sigma}{N}\sum_{i=1}^N
  \int_0^t \nabla\varphi(X_i(s))\cdot dW_i(s),
  \label{eq:def-martingale}
\end{equation}
is a continuous martingale.
Let $\psi\in C_c^{1}((0,T))$ be a temporal test function, so that
$\psi(0)=\psi(T)=0$.
Integration by parts for semimartingales gives
\[
  d\bigl(\psi(t)\,\ip{\mu_t^N}{\varphi}\bigr)
  =
  \dot\psi(t)\,\ip{\mu_t^N}{\varphi}\,dt
  +
  \psi(t)\,d\ip{\mu_t^N}{\varphi},
\]
and integrating over $[0,T]$ eliminates the boundary terms.
Substituting \eqref{eq:empirical-martingale} into this integrated identity and
rearranging gives \eqref{eq:particle-weak-exact}. Since $\psi$ and
$\nabla\varphi$ are bounded and the tests are fixed independently of the
trajectories, the stochastic integral is square-integrable and has mean zero.
\end{proof}

\subsection{Finite-Particle Regression Estimates}
\label{app:moment-residual}

\subsubsection{Well-Posedness and Moment Bounds}

\begin{lemma}[Well-posedness and uniform moment bounds]
  \label{lem:IPS_moments}
  Suppose Assumption~\ref{ass:particle-system} holds, and let
  $q\ge2$ satisfy
  \[
    \E\|X_1(0)\|^q<\infty.
  \]
  Then \eqref{eq:SDE} admits a unique strong solution and there exists $C_X>0$,
  independent of $N$, such that
  \[
    \sup_{N\in\N}\ \sup_{1\le i\le N}\ \sup_{t\in[0,T]}
    \E\|X_i(t)\|^q\le C_X.
  \]
\end{lemma}

The constant $C_X$ depends only on $T$, $q$, and $d$, together with the kernel
constants $L_K$ and $G_K$. It may also depend on $\sigma$ and the initial moment
$\E\|X_1(0)\|^q$.

\begin{proof}[Proof of Lemma~\ref{lem:IPS_moments}]
  We apply the classical strong-solution theorems of
  \citet[Theorems~5.2.5 and~5.2.9]{Karatzas1998Brownian} for each fixed
  $N$, then use the standard localized It\^o--Gr\"onwall argument to obtain
  componentwise moments uniformly in $N$.
  Fix $N\in\N$.
  For $\mathbf x=(x_1,\dots,x_N)\in(\R^d)^N$, write the interacting drift as
  \[
    v_i^N(\mathbf x)
    := \frac1N\sum_{j\neq i} K^\star(x_i-x_j)\in\R^d,
    \qquad i=1,\dots,N,
  \]
  and set
  \[
    v^N(\mathbf x):=\bigl(v_1^N(\mathbf x),\dots,v_N^N(\mathbf x)\bigr).
  \]
  Then the $Nd$-dimensional SDE reads
  \[
    dX(t)=v^N(X(t))\,dt+\sigma\,dW(t),
    \qquad X(t)=(X_1(t),\dots,X_N(t))\in(\R^d)^N,
  \]
  with $W(t)=(W_1(t),\dots,W_N(t))$ and independent $d$-dimensional Brownian motions.

  \paragraph{Step 1: Strong existence and uniqueness.}
  Assumption~\ref{ass:particle-kernel} gives, for
  $\mathbf x,\mathbf y\in(\R^d)^N$,
  \begin{align*}
    \|v_i^N(\mathbf x)-v_i^N(\mathbf y)\|
     &\le L_K\|x_i-y_i\|+\frac{L_K}{N}\sum_{j=1}^N\|x_j-y_j\|, \\
    \|v_i^N(\mathbf x)\|
     &\le G_K\Bigl(1+\|x_i\|
       +\frac1N\sum_{j=1}^N\|x_j\|\Bigr).
  \end{align*}
  Cauchy--Schwarz in the particle average yields
  $\|v^N(\mathbf x)-v^N(\mathbf y)\|\le2L_K\|\mathbf x-\mathbf y\|$.
  Thus the full drift is globally Lipschitz and has linear growth for each
  fixed $N$. The componentwise growth bound, whose constants are independent
  of $N$, will give the uniform moment estimate.

  For this fixed $N$, the initial vector has a finite second moment and is
  independent of $W$. The bounds on $v^N$ and the constant diffusion
  coefficient satisfy the Lipschitz and linear-growth conditions in
  \cite[Theorems~5.2.5 and~5.2.9]{Karatzas1998Brownian}. Hence the
  $Nd$-dimensional SDE admits a unique global strong solution on $[0,T]$.

  \paragraph{Step 2: Uniform moment bound.}
  The derivatives needed to apply It\^o's formula
  \cite{Karatzas1998Brownian} to $\|X_i(t)\|^q$ are
  \begin{equation}
    \nabla\|x\|^q=q\|x\|^{q-2}x,
    \qquad \Delta\|x\|^q=q(q+d-2)\|x\|^{q-2}.
    \label{eq:norm-q-derivatives}
  \end{equation}
  Hence, with $C_q:=q(q+d-2)$,
  \begin{equation}\label{eq:q-moment-ito}
    d\|X_i(t)\|^q
    =\|X_i(t)\|^{q-2}\Bigl[qX_i(t)\cdot v_i^N(X(t))\,dt
    +q\sigma X_i(t)\cdot dW_i(t)+\frac{\sigma^2}{2}C_q\,dt\Bigr].
  \end{equation}
  Define the local martingale
  \[
    M_i^{(q)}(t):= q\sigma\int_0^t \|X_i(s)\|^{q-2}X_i(s)\cdot dW_i(s),
  \]
  and, for $R>0$, the stopping time
  \[
    \tau_R := \inf\Bigl\{t\ge 0:\ \max_{1\le i\le N}\|X_i(t)\|\ge R\Bigr\}.
  \]
  On $[0,\tau_R]$ the integrand of $M_i^{(q)}$ is bounded, so the stopped process
  $M_i^{(q)}(\cdot\wedge\tau_R)$ is a true martingale and
  $\E[M_i^{(q)}(t\wedge\tau_R)]=0$.
  By Step~1 the solution is global, so $\tau_R\uparrow\infty$ almost surely as
  $R\to\infty$.
  Using $\|x\|^{q-2}\le 1+\|x\|^q$ for $q\ge2$, we obtain from \eqref{eq:q-moment-ito}
  \begin{equation}\label{eq:q-moment-ineq2}
    d\|X_i(t)\|^q
    \le q\|X_i(t)\|^{q-2}X_i(t)\cdot v_i^N(X(t))\,dt
    +C\,\bigl(1+\|X_i(t)\|^q\bigr)\,dt+dM_i^{(q)}(t),
  \end{equation}
  where $C$ depends only on $(q,\sigma,d)$.
  Young's inequality bounds the drift term by
  \[
    \|X_i(t)\|^{q-2}X_i(t)\cdot v_i^N(X(t))
    \le\frac{q-1}{q}\|X_i(t)\|^q+\frac1q\|v_i^N(X(t))\|^q.
  \]
  The componentwise growth bound and Jensen's inequality give
  \[
    \|v_i^N(\mathbf x)\|^q
    \le C\Bigl(1+\|x_i\|^q+\frac1N\sum_{j=1}^N\|x_j\|^q\Bigr),
  \]
  with $C$ depending only on $(q,G_K)$.
  Integrating \eqref{eq:q-moment-ineq2} over $[0,t\wedge\tau_R]$ and taking
  expectations (the stopped stochastic integral has zero mean), we find that the drift terms
  carry the indicator $\mathbf{1}_{\{s\le\tau_R\}}$. On this event,
  $X_j(s)=X_j(s\wedge\tau_R)$; bounding the indicator by one therefore controls
  the stopped integral by the stopped processes and yields
  \begin{equation}\label{eq:moment-diff-ineq}
    \E\|X_i(t\wedge\tau_R)\|^q
    \le \E\|X_i(0)\|^q+C\int_0^t
    \Bigl(1+\E\|X_i(s\wedge\tau_R)\|^q
    +\frac1N\sum_{j=1}^N\E\|X_j(s\wedge\tau_R)\|^q\Bigr)ds.
  \end{equation}
  Define the averaged, stopped $q$-moment
  \[
    m^R_q(t):=\frac1N\sum_{j=1}^N \E\|X_j(t\wedge\tau_R)\|^q,
  \]
  which is finite for every $R$, since the stopped trajectories are continuous and
  bounded by $\max(R,\|X_j(0)\|)$ up to time $\tau_R$.
  Averaging \eqref{eq:moment-diff-ineq} over $i=1,\dots,N$ yields
  \[
    m^R_q(t)\le m_q(0)+C\int_0^t \bigl(1+m^R_q(s)\bigr)\,ds,
    \qquad m_q(0):=\frac1N\sum_{j=1}^N\E\|X_j(0)\|^q.
  \]
  and Gr\"onwall's inequality implies, uniformly in $R$,
  \[
    m^R_q(t)\le \bigl(m_q(0)+1\bigr)e^{Ct}-1
    \le \bigl(\E\|X_1(0)\|^q+1\bigr)e^{CT}-1
    :=C_X,\qquad t\in[0,T],
  \]
  using i.i.d.\ initial data so that $m_q(0)=\E\|X_1(0)\|^q$.
  Letting $R\to\infty$ and applying Fatou's lemma (using
  $\tau_R\uparrow\infty$ and path continuity) gives
  \[
    m_q(t):=\frac1N\sum_{j=1}^N\E\|X_j(t)\|^q\ \le\ C_X,
    \qquad t\in[0,T].
  \]
  Finally, by permutation invariance of the system (i.i.d.\ initial conditions, independent
  Brownian motions, and symmetric interaction drift), the law of $(X_1(t),\dots,X_N(t))$
  is exchangeable for each $t$, hence $X_i(t)\stackrel{d}{=}X_1(t)$ for all $i$ and $t$.
  Therefore $\E\|X_i(t)\|^q = m_q(t)$ for all $i$, and thus
  \[
    \sup_{i=1,\dots,N}\sup_{t\in[0,T]}\E\|X_i(t)\|^q
    = \sup_{t\in[0,T]} m_q(t)
    \le C_X.
  \]
  Taking the supremum over $N$ completes the proof.
\end{proof}

\subsubsection{Bounds for Residuals and Pair Sampling}
\label{sec:residual-decomposition}

For the continuous-time response and design
\eqref{eq:def-b-cts}--\eqref{eq:def-A-cts}, the exact representation stated in
Section~\ref{sec:main-result} and the weak identity
\eqref{eq:particle-weak-exact}, stacked over the test pairs with
the $1/\sqrt L$ normalization, give
\begin{equation}
  b^{N,\mathrm{cts}}
  =A^{N,\mathrm{cts}}\theta^\star+\eta^{N,\mathrm{cts}},
  \label{eq:master-regression-cts}
\end{equation}
where the martingale residual $\eta^{N,\mathrm{cts}}\in\R^L$ has entries
\begin{equation}
  \eta^{N,\mathrm{cts}}_a
  :=\frac{\sigma}{N\sqrt L}\sum_{i=1}^N\int_0^T
    \psi_k(t)\,\nabla\varphi_\ell(X_i(t))\cdot dW_i(t).
  \label{eq:def-eta-stoch}
\end{equation}
For the latent discrete regression, define
\begin{equation}
  \eta^{N,\Delta t}:=b^{N,\Delta t}-A^{N,\Delta t}\theta^\star.
  \label{eq:master-regression-disc}
\end{equation}

Recall that $C_\varphi$ and $C_\psi$ are the test-function bounds from
Assumption~\ref{ass:test_functions_A}, $C_{\mathrm{quad}}$ is the quadrature
constant from Assumption~\ref{ass:time_grid_A}, and $\theta^\star$ is defined in
Section~\ref{sec:main-result}. Set
\begin{equation}
  C_\eta
  :=
  \max\left\{
  2T C_\varphi^2 C_\psi^2,\,
  4C_{\mathrm{quad}}\left(1+\|\theta^\star\|_2^2\right)
  \right\}.
  \label{eq:def-C-eta}
\end{equation}
The constant $C_\eta$ may depend on $d$ through
the fixed problem parameters, but is independent of $N$, $\Delta t$, and $L$.

\begin{lemma}[Residual second-moment bound]
  \label{lmm:residual-moment}
  Suppose Assumption~\ref{ass:particle-system}, part~(ii) of
  Assumption~\ref{ass:dictionary-model}, and
  Assumptions~\ref{ass:test_functions_A} and~\ref{ass:time_grid_A} hold. Then
  \begin{equation}
    \E\|\eta^{N,\Delta t}\|_2^2
    \le C_\eta\left(\frac{\sigma^2}{N}+(\Delta t)^{2p}\right).
    \label{eq:def-full-residual-scale}
  \end{equation}
\end{lemma}

\begin{proof}[Proof of Lemma~\ref{lmm:residual-moment}]
  Decompose
  \[
    \eta^{N,\Delta t}=\eta^{N,\mathrm{cts}}+\eta^{N,\mathrm{disc}},
    \qquad \eta^{N,\mathrm{disc}}:=b^{N,\Delta t}-b^{N,\mathrm{cts}}
    +(A^{N,\mathrm{cts}}-A^{N,\Delta t})\theta^\star.
  \]
  Since $\supp(\theta^\star)=S$, the matrix term depends only on the columns
  indexed by $S$.

  For the stochastic term, we follow the martingale argument of
  \citet[Lemma~2 and Appendix~A.4]{MessengerLearning2022a}, with the test
  normalization and constants of the present regression.
  For a row $a=(\ell,k)$, It\^o's isometry and the orthogonality of the stochastic
  integrals with respect to distinct Brownian motions
  \cite{Karatzas1998Brownian} give
  \[
    \E\bigl|\eta^{N,\mathrm{cts}}_{a}\bigr|^2
    =
    \frac{\sigma^2}{L N^2}
    \sum_{i=1}^N
    \E\int_0^T
    \psi_k(t)^2
    \|\nabla\varphi_\ell(X_i(t))\|^2\,dt.
  \]
  Since
  \[
    \|\psi_k\|_{L^2(0,T)}^2
    \le
    T\|\psi_k\|_{L^\infty(0,T)}^2
    \le
    T C_\psi^2,
  \]
  Assumption~\ref{ass:test_functions_A} yields
  \[
    \E\bigl|\eta^{N,\mathrm{cts}}_{a}\bigr|^2
    \le
    \frac{\sigma^2 C_\varphi^2 T C_\psi^2}{L N}.
  \]
  Summing over the $L=L_xL_t$ rows gives
  \begin{equation}
    \E\left[\bigl\|\eta^{N,\mathrm{cts}}\bigr\|_2^2\right]
    \le
    T C_\varphi^2 C_\psi^2\frac{\sigma^2}{N}.
    \label{eq:eta-stoch-vector-bound}
  \end{equation}

  For the discretization term, use
  $\|u+v\|_2^2\le2\|u\|_2^2+2\|v\|_2^2$ and
  $\|B\theta^\star\|_2\le\|B\|_F\|\theta^\star\|_2$.
  Assumption~\ref{ass:time_grid_A} then gives
  \begin{align}
    \E\|\eta^{N,\mathrm{disc}}\|_2^2
    &\le2\E\|b^{N,\Delta t}-b^{N,\mathrm{cts}}\|_2^2
    +2\|\theta^\star\|_2^2\E\|A^{N,\Delta t}-A^{N,\mathrm{cts}}\|_F^2 \notag\\
    &\le2C_{\mathrm{quad}}(1+\|\theta^\star\|_2^2)(\Delta t)^{2p}.
    \label{eq:eta-disc-bound}
  \end{align}

  Combining \eqref{eq:eta-stoch-vector-bound}--\eqref{eq:eta-disc-bound}
  with the two-term inequality gives
  \[
    \E\left[\bigl\|\eta^{N,\Delta t}\bigr\|_2^2\right]
    \le
    2T C_\varphi^2 C_\psi^2\frac{\sigma^2}{N}
    +
    4C_{\mathrm{quad}}\left(1+\|\theta^\star\|_2^2\right)(\Delta t)^{2p}.
  \]
  The definition \eqref{eq:def-C-eta} therefore proves the stated bound.
\end{proof}

At the true kernel, write
$\eta_Y^{N,\Delta t}:=\eta_Y^{N,\Delta t}(K^\star)$. Under exact realizability,
the regression identity in Section~\ref{sec:regression-representation} gives
$\eta_Y^{N,\Delta t}=b_Y^{N,\Delta t}-A_Y^{N,\Delta t}\theta^\star$.
To compare this residual with the latent regression, define the observation
perturbations
\begin{equation}
  E_b^{N,\Delta t}:=b_Y^{N,\Delta t}-b^{N,\Delta t},
  \qquad E_A^{N,\Delta t}:=A_Y^{N,\Delta t}-A^{N,\Delta t}.
  \label{eq:def-observation-perturbations}
\end{equation}
Subtracting the corresponding response--design identities gives
\begin{equation}
  \eta_Y^{N,\Delta t}
  =\eta^{N,\Delta t}+E_b^{N,\Delta t}-E_A^{N,\Delta t}\theta^\star.
  \label{eq:observed-residual-decomposition}
\end{equation}

\begin{lemma}[Observation-noise perturbation bounds]
  \label{lem:observation-perturbation}
  Suppose Assumptions~\ref{ass:particle-system} and
  \ref{ass:observation-noise}, part~(i) of
  Assumption~\ref{ass:dictionary-model}, and
  Assumptions~\ref{ass:test_functions_A} and~\ref{ass:time_grid_A} hold, with
  $r\ge3$ in Assumption~\ref{ass:test_functions_A}. Then there
  exist constants $C_b^{\mathrm{obs}},C_A^{\mathrm{obs}}>0$ such that
  \begin{align}
    \E\|E_b^{N,\Delta t}\|_2^2
     & \le C_b^{\mathrm{obs}}\sigma_{\mathrm{obs}}^2,
    \label{eq:observation-response-moment}\\
    \E\|E_A^{N,\Delta t}\|_F^2
     & \le C_A^{\mathrm{obs}}\sigma_{\mathrm{obs}}^2.
    \label{eq:observation-design-moment}
  \end{align}
  If part~(ii) of Assumption~\ref{ass:dictionary-model} also holds, then there exists
  $C_\eta^{\mathrm{obs}}>0$ such that
  \begin{equation}
    \E\|E_A^{N,\Delta t}\theta^\star\|_2^2
    \le C_\eta^{\mathrm{obs}}\sigma_{\mathrm{obs}}^2.
    \label{eq:observation-directional-moment}
  \end{equation}
  The constants are uniform in $N$, $L$, $\Delta t$, and
  $\sigma_{\mathrm{obs}}\in[0,\overline\sigma_{\mathrm{obs}}]$; they may depend on the
  fixed model, dictionary, test-function, and quadrature constants. In
  particular, $C_b^{\mathrm{obs}}$ may depend on the known process-noise
  intensity $\sigma$ through the It\^o-correction term. The directional
  constant does not depend explicitly on $s$ or $\theta_S^\star$.
\end{lemma}

\begin{proof}[Proof of Lemma~\ref{lem:observation-perturbation}]
  Write $\varepsilon_{i,n}:=\varepsilon^{\mathrm{obs}}_{i,n}$ within this proof.
  The mean-value theorem and Assumption~\ref{ass:test_functions_A} with
  $r\ge3$ give
  \begin{align}
    |\varphi_\ell(Y_i(t_n))-\varphi_\ell(X_i(t_n))|
     & \le C_\varphi\|\varepsilon_{i,n}\|,
    \label{eq:observation-phi-difference}\\
    |\Delta\varphi_\ell(Y_i(t_n))-\Delta\varphi_\ell(X_i(t_n))|
     & \le C\|\varepsilon_{i,n}\|.
    \label{eq:observation-laplacian-difference}
  \end{align}
  For any scalars $z_n$, the quadrature weights satisfy
  \begin{equation}
    \left|\sum_n w_nz_n\right|^2
    \le W_{\Delta t}\sum_n|w_n||z_n|^2.
    \label{eq:weighted-time-sum-bound}
  \end{equation}
  Applying Jensen's inequality to the particle average and then
  \eqref{eq:weighted-time-sum-bound}, we obtain from the first difference above,
  uniformly in $a=(\ell,k)$,
  \[
    \E\left|\frac1N\sum_i\sum_n w_n
    [\varphi_\ell(Y_i(t_n))-\varphi_\ell(X_i(t_n))]\dot\psi_k(t_n)\right|^2
    \le C\sigma_{\mathrm{obs}}^2.
  \]
  The same argument using \eqref{eq:observation-laplacian-difference} bounds
  the It\^o-correction difference by $C\sigma_{\mathrm{obs}}^2$. After the
  $1/\sqrt L$ row normalization, each component of $E_b^{N,\Delta t}$ has
  second moment at most $C\sigma_{\mathrm{obs}}^2/L$. Summing over the $L$ rows and
  combining the two response terms proves
  \eqref{eq:observation-response-moment}.

  For the design perturbation, fix $a=(\ell,k)$, $m$, and $i\neq j$, suppress
  the time arguments, and use the exact splitting
  \begin{align}
    &h_{a,m}(t_n,Y_i,Y_j)-h_{a,m}(t_n,X_i,X_j)
    =\psi_k(t_n)[\nabla\varphi_\ell(Y_i)-\nabla\varphi_\ell(X_i)]\cdot\Phi_m(X_i-X_j) \notag\\
    &\qquad+\psi_k(t_n)\nabla\varphi_\ell(Y_i)\cdot
    [\Phi_m(Y_i-Y_j)-\Phi_m(X_i-X_j)].
    \label{eq:observation-feature-splitting}
  \end{align}
  The uniform $C_b^r$ bound, the dictionary Lipschitz and linear-growth bounds,
  and the boundedness of $\psi_k$ imply that the absolute
  value of the first term is at most
  \[
    C\|\varepsilon_{i,n}\|
    \bigl(1+\|X_i(t_n)\|+\|X_j(t_n)\|\bigr),
  \]
  while the second is at most
  \[
    C\bigl(\|\varepsilon_{i,n}\|+\|\varepsilon_{j,n}\|\bigr).
  \]
  Independence of the observation errors from the latent particle system,
  Assumption~\ref{ass:observation-noise}, and the second-moment estimate of
  Lemma~\ref{lem:IPS_moments} therefore give
  \begin{equation}
    \E|h_{a,m}(t_n,Y_i,Y_j)-h_{a,m}(t_n,X_i,X_j)|^2
    \le C\sigma_{\mathrm{obs}}^2.
    \label{eq:observation-feature-difference-moment}
  \end{equation}
  Cauchy--Schwarz over the ordered pairs gives
  \[
    \left|\frac1{N^2}\sum_{i\neq j}z_{ij}\right|^2
    \le\frac{N(N-1)}{N^4}\sum_{i\neq j}|z_{ij}|^2.
  \]
  Combining this inequality with \eqref{eq:weighted-time-sum-bound} and
  \eqref{eq:observation-feature-difference-moment} shows that every entry of
  $E_A^{N,\Delta t}$ has second moment at most
  $C\sigma_{\mathrm{obs}}^2/L$. Summation over the $L$ rows and the fixed $M$
  dictionary atoms proves \eqref{eq:observation-design-moment}.

  For the directional estimate, realizability gives
  \[
    \sum_{m=1}^M\theta_m^\star h_{a,m}(t,x,y)
    =\psi_k(t)\nabla\varphi_\ell(x)\cdot K^\star(x-y).
  \]
  The difference of this expression at $(Y_i,Y_j)$ and $(X_i,X_j)$ admits the
  splitting \eqref{eq:observation-feature-splitting} with $\Phi_m$ replaced by
  $K^\star$. Assumption~\ref{ass:particle-kernel} and the same second-moment argument
  therefore give a bound $C\sigma_{\mathrm{obs}}^2$ for each directional pair
  difference. Repeating the pair and time averaging above and summing only over
  the $L$ rows proves \eqref{eq:observation-directional-moment}.
\end{proof}

\begin{remark}[Observation noise and temporal resolution]
  \label{rem:observation-noise-time-scaling}
  The bounds in Lemma~\ref{lem:observation-perturbation} contain no negative
  power of $\Delta t$. They use the stable total quadrature weight
  $W_{\Delta t}$ and do not rely on independence across observation times to
  obtain a $Q_{\Delta t}$ rate. By contrast, a forward difference contains
  \[
    \frac{Y_i(t_{n+1})-Y_i(t_n)}{\Delta t}
    =\frac{X_i(t_{n+1})-X_i(t_n)}{\Delta t}
    +\frac{\varepsilon^{\mathrm{obs}}_{i,n+1}
    -\varepsilon^{\mathrm{obs}}_{i,n}}{\Delta t}.
  \]
  Under Assumption~\ref{ass:observation-noise}, the second moment of the last
  term for each particle increment is
  $O(\sigma_{\mathrm{obs}}^2/(\Delta t)^2)$. This compares the scaling of the input
  perturbations before the respective regression normalizations and does not
  by itself assert that one coefficient estimator is uniformly more accurate
  than the other.
\end{remark}

For the sampled design, define the residual at the true coefficients by
\begin{equation}
  \widetilde\eta_Y^{N,\Delta t,P}
  :=b_Y^{N,\Delta t}-\widetilde A_Y^{N,\Delta t,P}\theta^\star.
  \label{eq:def-random-pair-residual}
\end{equation}
Since pair sampling leaves $b_Y^{N,\Delta t}$ unchanged,
\begin{equation}
  \widetilde\eta_Y^{N,\Delta t,P}
  =\eta_Y^{N,\Delta t}-D_Y^{N,\Delta t,P}\theta^\star.
  \label{eq:random-pair-residual-decomposition}
\end{equation}

\begin{proposition}[Observed residual and pair-sampling approximation bounds]
  \label{prop:residual-design-estimates}
  Suppose Assumptions~\ref{ass:particle-system},
  \ref{ass:observation-noise}, \ref{ass:dictionary-model},
  \ref{ass:test_functions_A}, and~\ref{ass:time_grid_A} hold, with $r\ge3$
  in Assumption~\ref{ass:test_functions_A}.
  There exist constants $C_\eta^Y,C_{\mathrm{pair},Y}^A,
  C_{\mathrm{pair},Y}^\eta>0$ such that, with the scale for $\eta_Y^{N,\Delta t}$
  \begin{equation}
    \mathcal V^Y_{N,\Delta t}
     :=C_\eta^Y\left(
       \frac{\sigma^2}{N}+(\Delta t)^{2p}+\sigma_{\mathrm{obs}}^2
     \right),
    \label{eq:def-observed-full-residual-scale}
  \end{equation}
  and the scale $\mathcal V^Y_{N,\Delta t,P}$ for $\widetilde\eta_Y^{N,\Delta t,P}$ defined in
  \eqref{eq:main-residual-scale}, the following estimates hold:
  \begin{enumerate}[label=(\roman*),leftmargin=2.2em]
    \item \textbf{Residual based on $A_Y^{N,\Delta t}$.} It satisfies
          \[
            \E\|\eta_Y^{N,\Delta t}\|_2^2
            \le \mathcal V^Y_{N,\Delta t}.
          \]
    \item \textbf{Residual based on $\widetilde A_Y^{N,\Delta t,P}$.} If
          Assumption~\ref{ass:random-pair-sampling} holds, then
          \[
            \E\|\widetilde\eta_Y^{N,\Delta t,P}\|_2^2
            \le \mathcal V^Y_{N,\Delta t,P}.
          \]
    \item \textbf{Pair-sampling perturbations.} If
          Assumption~\ref{ass:random-pair-sampling} holds, then
          \[
            \E[D_Y^{N,\Delta t,P}\mid\mathcal F_Y]=0,
            \quad \E\|D_Y^{N,\Delta t,P}\|_F^2
            \le C_{\mathrm{pair},Y}^A\frac{Q_{\Delta t}}{P},
            \quad \E\|D_Y^{N,\Delta t,P}\theta^\star\|_2^2
            \le C_{\mathrm{pair},Y}^\eta\frac{Q_{\Delta t}}{P}.
          \]
  \end{enumerate}
  The directional constant $C_{\mathrm{pair},Y}^\eta$ is obtained using the
  true kernel and does not depend explicitly on $s$ or $\theta_S^\star$.
\end{proposition}

\begin{proof}[Proof of Proposition~\ref{prop:residual-design-estimates}]
  \paragraph{Part~(i): observed full-pair residual.}
  The decomposition \eqref{eq:observed-residual-decomposition} and the
  three-term inequality give
  \[
    \E\|\eta_Y^{N,\Delta t}\|_2^2
    \le 3\E\|\eta^{N,\Delta t}\|_2^2
    +3\E\|E_b^{N,\Delta t}\|_2^2
    +3\E\|E_A^{N,\Delta t}\theta^\star\|_2^2.
  \]
  The residual estimate~\eqref{eq:def-full-residual-scale} from
  Lemma~\ref{lmm:residual-moment} and the perturbation bounds in
  Lemma~\ref{lem:observation-perturbation} bound the right-hand side by
  \[
    3C_\eta\left(\frac{\sigma^2}{N}+(\Delta t)^{2p}\right)
    +3\bigl(C_b^{\mathrm{obs}}+C_\eta^{\mathrm{obs}}\bigr)
    \sigma_{\mathrm{obs}}^2.
  \]
  Taking
  \[
    C_\eta^Y
    :=3\max\left\{C_\eta,
    C_b^{\mathrm{obs}}+C_\eta^{\mathrm{obs}}\right\}
  \]
  proves part~(i).

  \paragraph{Part~(iii): pair-sampling perturbations.}
  We condition on the observed trajectories and use the
  sampling-with-replacement calculation underlying the incomplete U-statistics
  of \citet[Section~3.1]{Clemencon2016Scalingup}.
  Independence of the pair draws across time then gives the sum of squared
  quadrature weights below.
  Fix a row $a=(\ell,k)$ and an atom $m$. For each time index $n$ define the
  observed full ordered-pair average
  \[
    \bar h_{Y,a,m,n}^N
    :=
    \frac{1}{N(N-1)}\sum_{i\neq j}
    h_{a,m}(t_n,Y_i(t_n),Y_j(t_n)).
  \]
  Since both $\widetilde A_Y^{N,\Delta t,P}$ and $A_Y^{N,\Delta t}$ carry the
  prefactor $1/\sqrt{L}$ and the correction $(N-1)/N$,
  \[
    \bigl(D_Y^{N,\Delta t,P}\bigr)_{a,m}
    =\frac{1}{\sqrt L}\frac{N-1}{N}\sum_n w_n\Biggl[
    \frac1P\sum_{\nu=1}^P
    h_{a,m}(t_n,Y_{I_{n,\nu}}(t_n),Y_{J_{n,\nu}}(t_n))
    -\bar h_{Y,a,m,n}^N\Biggr].
  \]
  By Assumption~\ref{ass:random-pair-sampling}, the conditional mean vanishes
  given $\mathcal F_Y$. Since the row and atom were arbitrary, this proves the
  conditional-centering claim in part~(iii). Independence of the pair samples
  across $n$ gives
  \begin{equation}
    \E\left[\left|\bigl(D_Y^{N,\Delta t,P}\bigr)_{a,m}\right|^2
    \bigm|\mathcal F_Y\right]
    \le\frac{1}{LP}\sum_n w_n^2\,
    \frac{1}{N(N-1)}\sum_{i\neq j}
    |h_{a,m}(t_n,Y_i(t_n),Y_j(t_n))|^2.
    \label{eq:random-pair-entry-conditional-variance}
  \end{equation}
  The observed positions have a uniform second-moment bound because
  \[
    \E\|Y_i(t_n)\|^2
    \le2\E\|X_i(t_n)\|^2
    +2C_\varepsilon\overline\sigma_{\mathrm{obs}}^2.
  \]
  The boundedness of $\psi_k$ and $\nabla\varphi_\ell$, together with the
  linear growth of $\Phi_m$, thus gives
  \[
    \E\left|\bigl(D_Y^{N,\Delta t,P}\bigr)_{a,m}\right|^2
    \le\frac{C Q_{\Delta t}}{LP}.
  \]
  Summing over the $L$ rows and fixed $M$ atoms gives
  \begin{equation}
    \E\|D_Y^{N,\Delta t,P}\|_F^2
    \le C_{\mathrm{pair},Y}^A\frac{Q_{\Delta t}}{P}.
    \label{eq:random-pair-design-moment}
  \end{equation}

  The indices $I_{n,\nu}$ and $J_{n,\nu}$ do not depend on $m$. Under the
  realizability assumption, this gives
  \[
    \sum_{m\in S}\theta_m^\star h_{a,m}(t_n,Y_i(t_n),Y_j(t_n))
    =\psi_k(t_n)\nabla\varphi_\ell(Y_i(t_n))\cdot K^\star(Y_i(t_n)-Y_j(t_n)).
  \]
  Repeating the conditional-variance calculation in
  \eqref{eq:random-pair-entry-conditional-variance} with $\Phi_m$ replaced by
  $K^\star$, and using its linear growth, gives
  \begin{equation}
    \E\|D_Y^{N,\Delta t,P}\theta^\star\|_2^2
    \le C_{\mathrm{pair},Y}^\eta\frac{Q_{\Delta t}}{P}.
    \label{eq:random-pair-parameter-moment}
  \end{equation}
  The estimate \eqref{eq:random-pair-parameter-moment} has no explicit
  dependence on $s$ or $\theta_S^\star$. Together with the conditional
  centering above, these are the three estimates in part~(iii).

  \paragraph{Part~(ii): sampled residual.}
  The residual $\eta_Y^{N,\Delta t}=b_Y^{N,\Delta t}
  -A_Y^{N,\Delta t}\theta^\star$ is $\mathcal F_Y$-measurable. Conditional
  centering therefore gives
  \begin{equation}
    \E\left\langle\eta_Y^{N,\Delta t},
    D_Y^{N,\Delta t,P}\theta^\star\right\rangle=0,
    \label{eq:random-pair-residual-orthogonality}
  \end{equation}
  Consequently, the decomposition
  \eqref{eq:random-pair-residual-decomposition} and the orthogonality relation
  \eqref{eq:random-pair-residual-orthogonality} give
  \begin{equation}
    \E\|\widetilde\eta_Y^{N,\Delta t,P}\|_2^2
    =\E\|\eta_Y^{N,\Delta t}\|_2^2
    +\E\|D_Y^{N,\Delta t,P}\theta^\star\|_2^2
    \le\mathcal V^Y_{N,\Delta t,P},
    \label{eq:random-pair-residual-moment}
  \end{equation}
  The estimate \eqref{eq:random-pair-residual-moment} proves part~(ii).
\end{proof}

Proposition~\ref{prop:residual-design-estimates} quantifies the weak-model
mismatch at the true kernel. The term
$\sigma^2/N$ is the finite-particle martingale contribution, while
$(\Delta t)^{2p}$ and $\sigma_{\mathrm{obs}}^2$ account for quadrature and
observation perturbations. Pair sampling adds a contribution proportional to
$Q_{\Delta t}/P$. Thus, even a realizable true kernel need not give zero
observed weak loss. Increasing $P$ reduces only the sampling contribution; it
does not remove the other sources of residual error. To turn these residual
bounds into coefficient bounds, we must also control the stability of the
design.

\subsubsection{Deterministic OLS and LASSO Bounds}

For a matrix $B$, let $B^\dagger$ denote its Moore--Penrose pseudoinverse.

\begin{proposition}[Oracle OLS coefficient error bound]
  \label{prop:generic-oracle}
  Let
  \[
    y=B\vartheta+\xi,
    \qquad
    \hat\vartheta:=B^\dagger y,
  \]
  where $B\in\R^{L\times s}$, $\vartheta\in\R^s$, and
  $\E\|\xi\|_2^2\le V$. Suppose that, for some $g>0$ and
  $\delta\in(0,1)$,
  \[
    \mathbb{P}\left(\sigma_{\min}(B)\ge g\right)\ge1-\delta.
  \]
  Let $\mathcal E_g:=\{\sigma_{\min}(B)\ge g\}$. Then:
  \begin{enumerate}[label=(\roman*),leftmargin=2.2em]
    \item for every $\varepsilon\in(0,1-\delta)$, with probability at least
          $1-\delta-\varepsilon$,
          \[
            \|\hat\vartheta-\vartheta\|_2^2
            \le\frac{V}{g^2\varepsilon};
          \]
    \item
          \[
            \E\left[
              \|\hat\vartheta-\vartheta\|_2^2
              \mathbf 1_{\mathcal E_g}
              \right]
            \le\frac{V}{g^2}.
          \]
  \end{enumerate}
\end{proposition}

\begin{proof}[Proof of Proposition~\ref{prop:generic-oracle}]
  We combine the standard pseudoinverse estimate with Markov's inequality.
  For $\varepsilon\in(0,1-\delta)$, let
  $\mathcal G_\varepsilon:=\{\|\xi\|_2^2\le V/\varepsilon\}$. Markov's
  inequality gives $\mathbb{P}(\mathcal G_\varepsilon^c)\le\varepsilon$, so the
  union bound gives
  $\mathbb{P}(\mathcal E_g\cap\mathcal G_\varepsilon)\ge1-\delta-\varepsilon$.
  On this event, $B$ has full column rank, so
  $\hat\vartheta-\vartheta=B^\dagger\xi$ and
  $\|B^\dagger\|_{\mathrm{op}}=1/\sigma_{\min}(B)\le1/g$. Hence
  \[
    \|\hat\vartheta-\vartheta\|_2^2
    \le\frac{\|\xi\|_2^2}{g^2}
    \le\frac{V}{g^2\varepsilon},
  \]
  proving (i). The same pseudoinverse bound on $\mathcal E_g$ gives
  \[
    \|\hat\vartheta-\vartheta\|_2^2\mathbf 1_{\mathcal E_g}
    \le\frac{\|\xi\|_2^2}{g^2};
  \]
  taking expectations proves (ii). The argument allows dependence between
  $B$ and $\xi$.
\end{proof}

For the LASSO bounds, use the error cone $\mathcal C_S$ from
Section~\ref{sec:main-result}. For a matrix
$B\in\R^{L\times M}$, write $B_{:,m}$ for its $m$th column and define
\begin{align}
  \gamma_{\mathrm{sp}}(B;S)
       & :=\inf_{u\in\mathcal C_S\setminus\{0\}}
  \frac{\|Bu\|_2}{\|u\|_2},
  \qquad B\in\R^{L\times M},
  \label{eq:def-cone-singular-value}                  \\
  R(B) & :=\max_{1\le m\le M}\|B_{:,m}\|_2.
  \label{eq:def-max-column-norm}
\end{align}
The quantity $\gamma_{\mathrm{sp}}(B;S)$ is a restricted-eigenvalue-type stability constant
under the normalization used here; see~\cite{Bickel2009Simultaneous,Geer2009conditions}. The quantity $R(B)$ is the
maximal column norm under the same normalization.
The main-text constant $R_\infty$ equals $R(A^\infty)$. When the true support $S$ is known, a positive
lower bound on $\sigma_{\min}(A_S^\infty)$ ensures that the active
coefficients are identifiable and controls the amplification of response
errors. For LASSO, a positive lower bound on
$\gamma_{\mathrm{sp}}(A^\infty;S)$ requires stability for all coefficient
changes in $\mathcal C_S$, including changes outside $S$.

For the generic regression below, define the score event
\begin{equation}
  \mathcal Z_\lambda(B,\xi)
  :=
  \left\{
  2\|B^\top\xi\|_\infty\le\lambda
  \right\}.
  \label{eq:def-lasso-score-event}
\end{equation}

\begin{proposition}[LASSO cone and coefficient error bound]
  \label{prop:generic-lasso}
  Fix $\lambda>0$. Let
  \[
    y=B\vartheta^\star+\xi,
    \qquad
    B\in\R^{L\times M},
  \]
  and let
  \[
    \hat\vartheta^\lambda
    \in
    \argmin_{\vartheta\in\R^M}
    \left\{
    \frac12\|B\vartheta-y\|_2^2
    +
    \lambda\|\vartheta\|_1
    \right\}.
  \]
  Let
  \[
    S:=\operatorname{supp}(\vartheta^\star),
    \qquad s:=|S|,
    \qquad
    u:=\hat\vartheta^\lambda-\vartheta^\star.
  \]
  Here $\mathcal C_S$ is the error cone from Section~\ref{sec:main-result}, and $\gamma_{\mathrm{sp}}(B;S)$ is defined by \eqref{eq:def-cone-singular-value} for this support $S$.
  On the event $\mathcal Z_\lambda(B,\xi)$,
  \[
    \|Bu\|_2^2
    +
    \lambda\|u_{S^c}\|_1
    \le
    3\lambda\|u_S\|_1.
  \]
  Consequently,
  \[
    u\in\mathcal C_S.
  \]
  If, in addition, $\gamma_{\mathrm{sp}}(B;S)>0$, then
  \[
    \|\hat\vartheta^\lambda-\vartheta^\star\|_2^2
    \le
    \frac{9s\lambda^2}
    {\gamma_{\mathrm{sp}}(B;S)^4}.
  \]
\end{proposition}

\begin{proof}[Proof of Proposition~\ref{prop:generic-lasso}]
  We follow the deterministic basic-inequality and cone argument of
  \citet[proof of Lemma~11.2]{Geer2009conditions}; see also
  \citet{Bickel2009Simultaneous}. With the present loss
  normalization, optimality gives the basic inequality
  \[
    \frac12\|Bu\|_2^2
    \le
    \langle B^\top\xi,u\rangle
    +
    \lambda
    \left(
    \|\vartheta^\star\|_1
    -
    \|\vartheta^\star+u\|_1
    \right).
  \]
  On $\mathcal Z_\lambda(B,\xi)$,
  \[
    \langle B^\top\xi,u\rangle
    \le
    \|B^\top\xi\|_\infty\|u\|_1
    \le
    \frac{\lambda}{2}\|u\|_1.
  \]
  Since $\vartheta^\star_{S^c}=0$,
  \[
    \|\vartheta^\star\|_1-\|\vartheta^\star+u\|_1
    \le
    \|u_S\|_1-\|u_{S^c}\|_1.
  \]
  Substituting these two bounds and multiplying by two gives
  \[
    \|Bu\|_2^2
    +
    \lambda\|u_{S^c}\|_1
    \le
    3\lambda\|u_S\|_1.
  \]
  The cone inclusion follows. If
  $\gamma_{\mathrm{sp}}(B;S)>0$, then
  \[
    \gamma_{\mathrm{sp}}(B;S)^2\|u\|_2^2
    \le
    \|Bu\|_2^2
    \le
    3\lambda\|u_S\|_1
    \le
    3\lambda\sqrt{s}\|u\|_2.
  \]
  If $u=0$ there is nothing to prove; otherwise divide by $\|u\|_2$
  and square the resulting bound.
\end{proof}

For the regression using $\widetilde A_Y^{N,\Delta t,P}$, Propositions~\ref{prop:generic-oracle}
and~\ref{prop:generic-lasso} apply with
$B=\widetilde A_Y^{N,\Delta t,P}$ and
$\xi=\widetilde\eta_Y^{N,\Delta t,P}$; oracle OLS uses the restriction
$B=\widetilde A_{Y,S}^{N,\Delta t,P}$.
Proposition~\ref{prop:residual-design-estimates} supplies the residual moments,
and Appendix~\ref{app:continuum} supplies the required high-probability design
stability events.

\subsection{Analysis of Finite-Particle Designs and Their Continuum Limit}
\label{app:mf-design}

We use the McKean--Vlasov process defined in
\eqref{eq:mckean-vlasov-sde} and its law $\rho_t$ from
Section~\ref{sec:main-result}. Coupling the interacting particle system with
independent copies bounds $A^{N,\mathrm{cts}}-A^\infty$ and
$A^{N,\Delta t}-A^\infty$.
We then control the additional perturbations due to observation errors and pair sampling.

\subsubsection{McKean--Vlasov Coupling and Feature Stability}

For a probability measure $\mu$ with finite first moment, write
\[
  v_\mu(x):=\int_{\R^d}K^\star(x-y)\,\mu(dy).
\]

\begin{lemma}[McKean--Vlasov moment bound]
  \label{lem:MV-moment}
  Suppose Assumption~\ref{ass:particle-system} holds and
  $\E\|X_1(0)\|^4<\infty$. Then the McKean--Vlasov SDE
  \eqref{eq:mckean-vlasov-sde} admits a unique strong solution $\bar X$, and
  there exists $C_{\mathrm{MV}}>0$ such that
  \[
    \sup_{t\in[0,T]}\E\|\bar X(t)\|^4\le C_{\mathrm{MV}}.
  \]
\end{lemma}

The constant $C_{\mathrm{MV}}$ depends only on $T$ and $d$, together with the
kernel constants $L_K$ and $G_K$. It may also depend on $\sigma$ and the initial
fourth moment $\E\|X_1(0)\|^4$.

\begin{proof}[Proof of Lemma~\ref{lem:MV-moment}]
  We apply the well-posedness result of
  \citet[Theorem~2.2 and its proof]{Meleard1996Asymptotic} to the two-variable
  drift $K^\star(x-y)$ and constant diffusion matrix $\sigma I_d$.
  By Assumption~\ref{ass:particle-kernel},
  \[
    \begin{aligned}
      \|v_\mu(x)-v_{\mu'}(x')\|
      &\le L_K\|x-x'\| + L_K\,\mathcal{W}_1(\mu,\mu'), \\
      \|v_\mu(x)\|
      &\le G_K\Bigl(1+\|x\|+\int\|y\|\,\mu(dy)\Bigr),
    \end{aligned}
  \]
  where $\mathcal{W}_1$ denotes the Wasserstein-$1$ distance.
  The two-variable coefficients are globally Lipschitz, and the assumed fourth
  moment implies a finite second moment. The construction in
  \cite[proof of Theorem~2.2]{Meleard1996Asymptotic} therefore yields a unique
  strong solution; see also \cite{Sznitman1991Topics,Chaintron2022Propagationa}.

  The same construction also gives
  $\sup_{t\in[0,T]}\E\|\bar X(t)\|^2<\infty$. The growth bound above therefore
  implies $\|v_{\rho_t}(x)\|\le C(1+\|x\|)$ uniformly in time.
  For the fourth moment, use \eqref{eq:norm-q-derivatives} with $q=4$ and the
  localized It\^o argument of Lemma~\ref{lem:IPS_moments}. With
  $\bar\tau_R:=\inf\{t\ge0:\|\bar X(t)\|\ge R\}$, it gives
  \[
    \E\|\bar X(t\wedge\bar\tau_R)\|^4
    \le\E\|\bar X(0)\|^4+C\int_0^t
    \bigl(1+\E\|\bar X(s\wedge\bar\tau_R)\|^4\bigr)\,ds,
  \]
  with $C$ independent of $R$. Gr\"onwall's inequality, followed by Fatou's
  lemma as $R\to\infty$, yields the stated bound with the claimed parameter
  dependence.
\end{proof}

For $i=1,\dots,N$, let $\bar X_i$ solve
\[
  d\bar X_i(t)=(K^\star*\rho_t)(\bar X_i(t))\,dt+\sigma\,dW_i(t),
  \qquad \bar X_i(0)=X_i(0),
\]
using the same initial condition and Brownian motion as $X_i$. These are the
i.i.d.\ nonlinear copies in the synchronous coupling.

\begin{lemma}[Strong propagation-of-chaos estimate]
  \label{lem:strong-poc}
  Suppose Assumption~\ref{ass:particle-system} holds and
  $\E\|X_1(0)\|^4<\infty$. Under the synchronous coupling, there exists
  $C_{\mathrm{pc}}>0$ such that
  \[
    \sup_{1\le i\le N}\ \sup_{t\in[0,T]}
    \E\|X_i(t)-\bar X_i(t)\|^4
    \le\frac{C_{\mathrm{pc}}}{N^2}.
  \]
\end{lemma}

The constant $C_{\mathrm{pc}}$ depends only on $T$ and $d$, together with the
kernel constants $L_K$ and $G_K$. It may also depend on $\sigma$ and the initial
fourth moment $\E\|X_1(0)\|^4$. By Cauchy--Schwarz,
\[
  \sup_{1\le i\le N}\ \sup_{t\in[0,T]}
  \E\|X_i(t)-\bar X_i(t)\|^2
  \le C_{\mathrm{pc}}^{1/2}N^{-1}.
\]

\begin{proof}[Proof of Lemma~\ref{lem:strong-poc}]
  The proof follows the classical synchronous-coupling approach for globally
  Lipschitz McKean--Vlasov systems
  \cite{Sznitman1991Topics,Chaintron2022Propagationa}. We adapt the fourth-moment
  argument in \citet[proof of Lemma~5.2]{Meleard1996Asymptotic}
  to the present initial fourth-moment assumption and omission of
  self-interactions. The conditional fluctuation calculation below verifies
  the rate needed for the weighted feature estimate under these assumptions.
  Throughout the proof, $C$ denotes a constant whose value may change from line
  to line. It depends only on $T$ and $d$ together with the kernel constants
  $L_K$ and $G_K$. It may also depend on the noise level $\sigma$ and the initial
  fourth moment $\E\|X_1(0)\|^4$.
  By parts~(i)--(ii) of Assumption~\ref{ass:particle-system}, the pairs
  $(X_i(0),W_i)$ are i.i.d.\ across $i$. Since the coefficients of
  \eqref{eq:mckean-vlasov-sde} depend on
  the deterministic law $\rho_t$, each $\bar X_i$ is a measurable functional of
  $(X_i(0),W_i)$. Thus $\bar X_1,\dots,\bar X_N$ are i.i.d.
  McKean--Vlasov copies with time marginals $\rho_t$. By construction, each
  $X_i$ and $\bar X_i$ share the same initial condition and Brownian motion;
  this is the synchronous coupling.
  Write
  \[
    v_t(x):=v_{\rho_t}(x)=(K^\star*\rho_t)(x),
  \]
  and note the linear-growth bound
  \[
    \|v_t(x)\|
    \le
    G_K\bigl(1+\|x\|+\E\|\bar X(t)\|\bigr)
    \le
    C(1+\|x\|),
  \]
  by Lemma~\ref{lem:MV-moment}.

  \paragraph{Step 1: Difference dynamics.}
  Set $e_i(t):=X_i(t)-\bar X_i(t)$, so $e_i(0)=0$.
  Because $X_i$ and $\bar X_i$ are driven by the same Brownian motion, the stochastic
  terms cancel and $t\mapsto e_i(t)$ is absolutely continuous with
  \[
    \dot e_i(t)
    =
    \frac1N\sum_{j\neq i}
    \bigl[K^\star(X_i-X_j)-K^\star(\bar X_i-\bar X_j)\bigr]
    +
    \zeta_i(t),
  \]
  where
  \[
    \zeta_i(t):=\frac1N\sum_{j\neq i}K^\star(\bar X_i-\bar X_j)-v_t(\bar X_i)
    =\frac1N\sum_{j\neq i}Z_{ij}(t)-\frac1N\,v_t(\bar X_i).
  \]
  Here $Z_{ij}(t):=K^\star(\bar X_i(t)-\bar X_j(t))-v_t(\bar X_i(t))$; time arguments are
  suppressed when unambiguous.
  Note that the drift of the particle system uses the prefactor $1/N$ over the $N-1$
  terms $j\neq i$, which is why the deterministic self-term
  $-\tfrac1N v_t(\bar X_i)$ appears in the decomposition of $\zeta_i$.

  \paragraph{Step 2: Moments of the fluctuation term.}
  Fix $t$ and $i$, and condition on $\bar X_i(t)$.
  Since $\{\bar X_j(t)\}_{j\neq i}$ are i.i.d.\ with law $\rho_t$ and independent of
  $\bar X_i(t)$, the vectors $\{Z_{ij}(t)\}_{j\neq i}$ are conditionally i.i.d.\ and
  conditionally centered:
  \[
    \E\bigl[K^\star(\bar X_i-\bar X_j)\,\big|\,\bar X_i\bigr]
    =
    \int K^\star(\bar X_i-y)\,\rho_t(dy)
    =
    v_t(\bar X_i).
  \]
  By the linear growth of $K^\star$ and $v_t$ and the fourth-moment bound of
  Lemma~\ref{lem:MV-moment},
  \[
    \sup_{i\neq j}\ \sup_{t\in[0,T]}\ \E\|Z_{ij}(t)\|^4
    \le
    C .
  \]
  Expanding the fourth power and using conditional independence and centering,
  for each coordinate $c=1,\ldots,d$ we obtain
  \[
    \E\!\left[\left(\sum_{j\neq i}Z_{ij}^{(c)}\right)^4\,\middle|\,\bar X_i\right]
    =(N-1)\E\!\left[\left(Z_{ij}^{(c)}\right)^4\,\middle|\,\bar X_i\right]
    +3(N-1)(N-2)\left(\E\!\left[\left(Z_{ij}^{(c)}\right)^2\,\middle|\,\bar X_i\right]\right)^2.
  \]
  Using
  \[
    \|z\|^4
    \le
    d\sum_{c=1}^d |z^{(c)}|^4
  \]
  and summing over the coordinates, we obtain
  \[
    \E\!\left[\left\|\sum_{j\neq i}Z_{ij}\right\|^4\,\middle|\,\bar X_i\right]
    \le C\Bigl[(N-1)\E\!\left[\|Z_{ij}\|^4\mid\bar X_i\right]
    +(N-1)^2\left(\E\!\left[\|Z_{ij}\|^2\mid\bar X_i\right]\right)^2\Bigr].
  \]
  Taking expectations in the preceding conditional bound and dividing by
  $N^4$ gives
  \[
    \E\Bigl\|\frac1N\sum_{j\neq i}Z_{ij}\Bigr\|^4
    \le\frac{C}{N^4}\Bigl\{(N-1)\E\|Z_{ij}\|^4
    +(N-1)^2\E\!\left[\left(\E[\|Z_{ij}\|^2\mid\bar X_i]\right)^2\right]\Bigr\}.
  \]
  Jensen's inequality and the uniform fourth-moment estimate above give
  \[
    \E\left[(\E[\|Z_{ij}\|^2\mid\bar X_i])^2\right]
    \le\E\|Z_{ij}\|^4\le C.
  \]
  Therefore, after absorbing these moment bounds into $C$,
  \[
    \E\Bigl\|\frac1N\sum_{j\neq i}Z_{ij}\Bigr\|^4
    \le
    \frac{C}{N^4}\bigl[(N-1)+(N-1)^2\bigr]
    \le
    \frac{C}{N^2}.
  \]
  Moreover, the linear-growth bound for $v_t$ and
  Lemma~\ref{lem:MV-moment} give
  \[
    \E\|N^{-1}v_t(\bar X_i)\|^4
    =N^{-4}\E\|v_t(\bar X_i)\|^4
    \le\frac{C}{N^4}(1+\E\|\bar X_i\|^4)\le\frac{C}{N^4}.
  \]
  Hence
  \begin{equation}
    \sup_{1\le i\le N}\ \sup_{t\in[0,T]}\ \E\|\zeta_i(t)\|^4
    \le
    \frac{C}{N^2}.
    \label{eq:poc-fluctuation-fourth-moment}
  \end{equation}

  \paragraph{Step 3: Gr\"onwall estimate for the fourth moment.}
  By Assumption~\ref{ass:particle-kernel},
  \[
    \Bigl\|\frac1N\sum_{j\neq i}
    \bigl[K^\star(X_i-X_j)-K^\star(\bar X_i-\bar X_j)\bigr]\Bigr\|
    \le\frac{L_K}{N}\sum_{j\neq i}\bigl(\|e_i\|+\|e_j\|\bigr)
    \le L_K\Bigl(\|e_i\|+\frac1N\sum_{j=1}^N\|e_j\|\Bigr).
  \]
  Since $t\mapsto e_i(t)$ is absolutely continuous and $z\mapsto\|z\|^4$ is
  continuously differentiable,
  \[
    \frac{d}{dt}\|e_i\|^4=4\|e_i\|^2\,e_i\cdot\dot e_i
    \le4L_K\|e_i\|^4+4L_K\|e_i\|^3\cdot\frac1N\sum_{j=1}^N\|e_j\|
    +4\|e_i\|^3\|\zeta_i\|.
  \]
  Young's inequality $a^3b\le\tfrac34 a^4+\tfrac14 b^4$ and Jensen's inequality
  $\bigl(\tfrac1N\sum_j\|e_j\|\bigr)^4\le\tfrac1N\sum_j\|e_j\|^4$ give
  \[
    \frac{d}{dt}\|e_i\|^4
    \le
    C\|e_i\|^4
    +\frac{C}{N}\sum_{j=1}^N\|e_j\|^4
    +\|\zeta_i\|^4 .
  \]
  Integrating from $0$ to $t$ (recall $e_i(0)=0$), taking expectations, and
  averaging over the particle errors, set
  $D_4(t):=\frac1N\sum_{i=1}^N\E\|e_i(t)\|^4$. Using
  \eqref{eq:poc-fluctuation-fourth-moment}, we obtain
  \[
    D_4(t)
    \le
    C\int_0^t D_4(s)\,ds
    +\frac{C}{N^2},
    \qquad t\in[0,T].
  \]
  Since $D_4(0)=0$, Gr\"onwall's inequality yields
  $\sup_{t\in[0,T]}D_4(t)\le C_{\mathrm{pc}}/N^2$.

  Finally, the coupled system $\{(X_i,\bar X_i)\}_{i=1}^N$ is exchangeable (i.i.d.\ initial
  conditions, i.i.d.\ Brownian motions, and a drift that is symmetric in the particle
  indices). This exchangeability transfers the averaged estimate $D_4(t)$ to every
  particle: $\E\|e_i(t)\|^4=D_4(t)$ for every $i$, proving the fourth-moment
  bound; the second-moment bound follows from the Cauchy--Schwarz inequality.
\end{proof}

For $x,y,x',y'\in\R^d$, write
\[
  \omega(x,y,x',y')
  :=1+\|x\|+\|y\|+\|x'\|+\|y'\|.
\]

\begin{lemma}[Feature growth and coupling stability]
  \label{lem:feature-stability}
  Suppose part~(i) of Assumption~\ref{ass:dictionary-model},
  Assumption~\ref{ass:test_functions_A}, and the assumptions of
  Lemma~\ref{lem:strong-poc} hold. Then there exists $C_h>0$
  such that, uniformly in $a,m,t$:
  \begin{enumerate}[label=(\roman*),leftmargin=2.2em]
    \item
          \[
            |h_{a,m}(t,x,y)|\le C_h(1+\|x\|+\|y\|),
          \]
          and
          \[
            |h_{a,m}(t,x,y)-h_{a,m}(t,x',y')|
            \le C_h\omega(x,y,x',y')(\|x-x'\|+\|y-y'\|).
          \]
    \item for $i\ne j$,
          \[
            \E\left|
            h_{a,m}(t,X_i(t),X_j(t))
            -h_{a,m}(t,\bar X_i(t),\bar X_j(t))
            \right|^2
            \le\frac{C_h}{N}.
          \]
  \end{enumerate}
\end{lemma}

The constant $C_h$ depends only on $d$ and the uniform dictionary and
test-function bounds, together with the moment and coupling constants supplied
by Lemmas~\ref{lem:IPS_moments}, \ref{lem:MV-moment},
and~\ref{lem:strong-poc}. It is uniform in $N$ and time, as
well as in all particle and feature indices.

\begin{proof}[Proof of Lemma~\ref{lem:feature-stability}]
  Fix $a=(\ell,k)$ and $m$; all bounds below are uniform in these indices.
  By Assumption~\ref{ass:test_functions_A}, $|\psi_k(t)|\le C_\psi$,
  $\|\nabla\varphi_\ell\|_{L^\infty}\le C_\varphi$, and
  $\|\nabla^2\varphi_\ell\|_{L^\infty}\le C_\varphi$. By
  part~(i) of Assumption~\ref{ass:dictionary-model},
  $\|\Phi_m(z)\|\le G_\Phi(1+\|z\|)$ and $\Phi_m$ is $L_\Phi$-Lipschitz. The growth bound follows:
  \[
    |h_{a,m}(t,x,y)|
    \le
    C_\psi C_\varphi G_\Phi\bigl(1+\|x-y\|\bigr)
    \le
    C\,(1+\|x\|+\|y\|).
  \]
  For the stability bound, decompose
  \begin{align}
    &\nabla\varphi_\ell(x)\cdot\Phi_m(x-y)
    -\nabla\varphi_\ell(x')\cdot\Phi_m(x'-y')
    =\bigl[\nabla\varphi_\ell(x)-\nabla\varphi_\ell(x')\bigr]\cdot\Phi_m(x-y) \\
    &\qquad+\nabla\varphi_\ell(x')\cdot\bigl[\Phi_m(x-y)-\Phi_m(x'-y')\bigr].
  \end{align}
  By the uniform $C_b^r$ bound, $\nabla\varphi_\ell$ is
  $C_\varphi$-Lipschitz, so the first term is bounded in absolute value by
  \[
    C_\varphi\,\|x-x'\|\cdot G_\Phi\bigl(1+\|x\|+\|y\|\bigr),
  \]
  while the second term is bounded by
  $C_\varphi L_\Phi\bigl(\|x-x'\|+\|y-y'\|\bigr)$.
  Multiplying by $|\psi_k(t)|\le C_\psi$ and combining proves the stability bound.

  For the last claim, apply the stability bound with
  $(x,y)=(X_i(t),X_j(t))$ and $(x',y')=(\bar X_i(t),\bar X_j(t))$, and set
  \[
    \omega_t:=\omega(X_i,X_j,\bar X_i,\bar X_j),
    \qquad \Xi_{ij}(t):=\|X_i-\bar X_i\|+\|X_j-\bar X_j\|.
  \]
  By the Cauchy--Schwarz inequality,
  \[
    \E\bigl[|h_{a,m}(t,X_i,X_j)-h_{a,m}(t,\bar X_i,\bar X_j)|^2\bigr]
    \le C\,\E\bigl[\omega_t^2\Xi_{ij}(t)^2\bigr]
    \le C\,\bigl(\E\omega_t^4\bigr)^{1/2}\bigl(\E\Xi_{ij}(t)^4\bigr)^{1/2}.
  \]
  Lemma~\ref{lem:IPS_moments} with $q=4$ and Lemma~\ref{lem:MV-moment} give
  $\E \omega_t^4\le C$, while Lemma~\ref{lem:strong-poc} gives
  $\E \Xi_{ij}(t)^4\le C/N^2$. Hence the left-hand side is bounded by $C/N$, uniformly in
  $i\neq j$, $t$, $a$, and $m$.
\end{proof}

\subsubsection{Design Approximation Bounds}

\begin{proposition}[Bounds for finite-particle designs relative to the continuum]
  \label{prop:weak-design-convergence}
  Suppose Assumption~\ref{ass:particle-system}, part~(i) of
  Assumption~\ref{ass:dictionary-model}, and
  Assumptions~\ref{ass:test_functions_A} and~\ref{ass:time_grid_A} hold, and
  assume in addition that
  $\E\|X_1(0)\|^4<\infty$. There exist $C_A,C_A^Y>0$ such that, with
  the scales for matrices using all ordered pairs
  \begin{align}
    \mathcal D_{N,\Delta t}
     & :=C_A\left(\frac1N+(\Delta t)^{2p}\right),
    \label{eq:def-full-design-scale}\\
    \mathcal D^Y_{N,\Delta t}
     & :=C_A^Y\left(
       \frac1N+(\Delta t)^{2p}+\sigma_{\mathrm{obs}}^2
     \right),
    \label{eq:def-observed-full-design-scale}
  \end{align}
  and the sampled design scale $\mathcal D^Y_{N,\Delta t,P}$ defined in
  \eqref{eq:main-design-scale}, the following estimates hold:
  \begin{enumerate}[label=(\roman*),leftmargin=2.2em]
    \item The latent designs satisfy
          \[
            \E\|A^{N,\mathrm{cts}}-A^\infty\|_F^2\le\frac{C_A}{N},
            \qquad
            \E\|A^{N,\Delta t}-A^\infty\|_F^2
            \le \mathcal D_{N,\Delta t}.
          \]
    \item If Assumption~\ref{ass:observation-noise} also holds and
          Assumption~\ref{ass:test_functions_A} holds with $r\ge3$, then
          \[
            \E\|A_Y^{N,\Delta t}-A^\infty\|_F^2
            \le \mathcal D^Y_{N,\Delta t}.
          \]
    \item If, in addition, Assumption~\ref{ass:random-pair-sampling} holds,
          then
          \[
            \E\|\widetilde A_Y^{N,\Delta t,P}-A^\infty\|_F^2
            \le \mathcal D^Y_{N,\Delta t,P}.
          \]
  \end{enumerate}
  The same estimates hold with the Frobenius norm on each left-hand side
  replaced by the operator norm.
\end{proposition}

\begin{proof}[Proof of Proposition~\ref{prop:weak-design-convergence}]
  \paragraph{Part~(i): latent designs.}
  For $N=1$, the continuous-time bound follows from $A^{1,\mathrm{cts}}=0$
  and the feature moment bounds. We take $N\ge2$ in Steps~1--2; the quadrature
  argument in Step~3 applies to every $N$.
  Under the stated assumptions, Lemma~\ref{lem:strong-poc} applies. Let
  $\{\bar X_i\}_{i=1}^N$ be its coupled nonlinear copies, and define the auxiliary
  i.i.d.-copy design matrix
  \[
    A^{N,\mathrm{iid}}_{a,m}
    :=
    \frac{1}{\sqrt L}\,
    \frac{1}{N^2}\sum_{i\neq j}
    \int_0^T h_{a,m}\bigl(t,\bar X_i(t),\bar X_j(t)\bigr)\,dt .
  \]
  We first prove the continuous-time estimate. The comparison proceeds in two
  stages: the coupling replaces the interacting trajectories $X_i$ by the
  independent copies $\bar X_i$, and i.i.d.\ empirical averaging then replaces
  the finite pair average by its population counterpart. Accordingly, we use
  the splitting
  \begin{equation}
    A^{N,\mathrm{cts}}-A^\infty
    =
    \bigl(A^{N,\mathrm{cts}}-A^{N,\mathrm{iid}}\bigr)
    +
    \bigl(A^{N,\mathrm{iid}}-A^\infty\bigr).
    \label{eq:continuous-design-splitting}
  \end{equation}
  We estimate these two terms entrywise, sum over the $L\times M$ entries, and
  add the time-discretization error at the end.

  \paragraph{Step 1: Coupling part.}
  We apply Lemma~\ref{lem:feature-stability} to transfer the coupling estimate of
  Lemma~\ref{lem:strong-poc} to the features. We average these feature
  differences over the ordered pairs and time. For each entry, write
  \[
    \begin{aligned}
      A^{N,\mathrm{cts}}_{a,m}-A^{N,\mathrm{iid}}_{a,m}
      &=\frac{1}{\sqrt L}\int_0^T
      \frac{1}{N^2}\sum_{i\neq j} \delta h^{a,m}_{ij}(t)\,dt, \\
      \delta h^{a,m}_{ij}(t)
      &:=h_{a,m}(t,X_i,X_j)-h_{a,m}(t,\bar X_i,\bar X_j).
    \end{aligned}
  \]
  By the Cauchy--Schwarz inequality in time,
  $|\int_0^T f(t)\,dt|^2\le T\int_0^T|f(t)|^2\,dt$, and by the Cauchy--Schwarz
  inequality over the $N(N-1)$ ordered pairs,
  \[
    \Bigl|\frac{1}{N^2}\sum_{i\neq j}\delta h^{a,m}_{ij}(t)\Bigr|^2
    \le
    \frac{N(N-1)}{N^4}\sum_{i\neq j}\bigl|\delta h^{a,m}_{ij}(t)\bigr|^2 .
  \]
  Taking expectations and using Lemma~\ref{lem:feature-stability},
  $\E|\delta h^{a,m}_{ij}(t)|^2\le C/N$, we get
  \[
    \E\,\Bigl|\frac{1}{N^2}\sum_{i\neq j}\delta h^{a,m}_{ij}(t)\Bigr|^2
    \le
    \frac{N(N-1)}{N^4}\cdot N(N-1)\cdot\frac{C}{N}
    \le
    \frac{C}{N},
  \]
  hence
  \[
    \E\bigl|A^{N,\mathrm{cts}}_{a,m}-A^{N,\mathrm{iid}}_{a,m}\bigr|^2
    \le
    \frac{T}{L}\int_0^T \frac{C}{N}\,dt
    =
    \frac{CT^2}{LN}
    \le
    \frac{C}{LN}.
  \]
  In the last inequality, the fixed factor $T^2$ is absorbed into $C$.
  Summing over the $L$ rows and all $M$ dictionary columns,
  \begin{equation}
    \E\bigl\|A^{N,\mathrm{cts}}-A^{N,\mathrm{iid}}\bigr\|_F^2
    \le
    \frac{CM}{N}.
    \label{eq:continuous-design-coupling-bound}
  \end{equation}

  \paragraph{Step 2: i.i.d.\ part (ordered-pair U-statistic).}
  We apply Hoeffding's variance bound to the independent copies after
  symmetrizing the path kernel. For fixed $(a,m)$, define
  \[
    G^{a,m}_{ij}
    :=
    \int_0^T h_{a,m}\bigl(t,\bar X_i(t),\bar X_j(t)\bigr)\,dt,
    \qquad i\neq j .
  \]
  By the growth bound of Lemma~\ref{lem:feature-stability}, the Cauchy--Schwarz
  inequality in time, and the moment bound of Lemma~\ref{lem:MV-moment},
  \[
    \sup_{i\neq j}\ \sup_{a,m}\ \E\bigl|G^{a,m}_{ij}\bigr|^2
    \le
    T\int_0^T \E\bigl|h_{a,m}(t,\bar X_i,\bar X_j)\bigr|^2\,dt
    \le
    C .
  \]
  Since the paths $\bar X_1,\dots,\bar X_N$ are i.i.d., the variables $G^{a,m}_{ij}$,
  $i\neq j$, are identically distributed; set $\bar G_{a,m}:=\E[G^{a,m}_{12}]$.
  By \eqref{eq:main-continuum-design}, the independence of
  $\bar X_1(t)$ and $\bar X_2(t)$, and Fubini's theorem,
  \[
    A^\infty_{a,m}
    =
    \frac{\bar G_{a,m}}{\sqrt L}.
  \]
  Define the ordered-pair statistic
  \[
    U^{a,m}_N
    :=
    \frac{1}{N^2}\sum_{i\neq j} G^{a,m}_{ij},
    \qquad\text{so that}\qquad
    A^{N,\mathrm{iid}}_{a,m}=\frac{U^{a,m}_N}{\sqrt L}.
  \]
  Since the sum runs over the $N(N-1)$ ordered pairs but is normalized by $N^2$,
  \[
    \E\bigl[U^{a,m}_N\bigr]
    =
    \frac{N(N-1)}{N^2}\,\bar G_{a,m}
    =
    \Bigl(1-\frac1N\Bigr)\bar G_{a,m},
  \]
  so the exclusion of the diagonal terms creates a bias
  $|\E[U^{a,m}_N]-\bar G_{a,m}|=|\bar G_{a,m}|/N$, whose square is bounded by $C/N^2$.
  The statistic $\frac{N}{N-1}U^{a,m}_N$ is a degree-two U-statistic with
  symmetrized path kernel $(G^{a,m}_{ij}+G^{a,m}_{ji})/2$. Its kernel has a
  bounded second moment by the preceding estimate. The variance
  inequality of \citet[Theorem~5.2, Eq.~(5.20)]{Hoeffding1948Class} therefore gives
  \[
    \mathrm{Var}(U^{a,m}_N)
    \le\left(\frac{N-1}{N}\right)^2\frac{2}{N}
    \mathrm{Var}\!\left(\frac{G^{a,m}_{12}+G^{a,m}_{21}}{2}\right)
    \le\frac{C}{N}.
  \]
  See also \cite{Serfling1980Approximation}. Combining this with the squared
  bias yields
  \[
    \E\bigl|U^{a,m}_N-\bar G_{a,m}\bigr|^2
    =\mathrm{Var}\bigl(U^{a,m}_N\bigr)
    +\frac{|\bar G_{a,m}|^2}{N^2}
    \le\frac{C}{N}.
  \]
  After the $1/\sqrt L$ normalization and summation over the $L\times M$
  entries,
  \begin{equation}
    \E\bigl\|A^{N,\mathrm{iid}}-A^\infty\bigr\|_F^2
    \le
    \frac{CM}{N}.
    \label{eq:continuous-design-iid-bound}
  \end{equation}

  \paragraph{Step 3: Continuous and discrete bounds.}
  In Steps~1--2, we bounded the two terms in the continuous-time splitting
  \eqref{eq:continuous-design-splitting}. Combining
  \eqref{eq:continuous-design-coupling-bound} and
  \eqref{eq:continuous-design-iid-bound} using
  $\|B+B'\|_F^2\le2\|B\|_F^2+2\|B'\|_F^2$ and absorbing $M$ into the
  constant gives the continuous-time Frobenius bound. The corresponding
  operator-norm bound follows from $\|B\|_{\mathrm{op}}\le\|B\|_F$.

  It remains to compare $A^{N,\Delta t}$ with $A^{N,\mathrm{cts}}$.
  Assumption~\ref{ass:time_grid_A} gives
  \[
    \E\bigl\|A^{N,\Delta t}-A^{N,\mathrm{cts}}\bigr\|_F^2
    \le C_{\mathrm{quad}}\,(\Delta t)^{2p}.
  \]
  Combining the continuous-time and quadrature estimates via
  $\|B+B'\|_F^2\le2\|B\|_F^2+2\|B'\|_F^2$ proves the Frobenius bound; the
  operator-norm bound follows from $\|B\|_{\mathrm{op}}\le\|B\|_F$.

  \paragraph{Part~(ii): observed full-pair design.}
  By the definition of $E_A^{N,\Delta t}$ in
  \eqref{eq:def-observation-perturbations},
  \[
    A_Y^{N,\Delta t}-A^\infty
    =E_A^{N,\Delta t}+\bigl(A^{N,\Delta t}-A^\infty\bigr).
  \]
  The two-term inequality, part~(i), and
  Lemma~\ref{lem:observation-perturbation} give
  \[
    \E\|A_Y^{N,\Delta t}-A^\infty\|_F^2
    \le2\mathcal D_{N,\Delta t}
    +2C_A^{\mathrm{obs}}\sigma_{\mathrm{obs}}^2.
  \]
  Taking
  \[
    C_A^Y:=2\max\{C_A,C_A^{\mathrm{obs}}\}
  \]
  proves the stated bound for $A_Y^{N,\Delta t}$. The operator-norm estimate follows
  from $\|B\|_{\mathrm{op}}\le\|B\|_F$.

  \paragraph{Part~(iii): sampled design.}
  We now separate the pair-sampling error from $A_Y^{N,\Delta t}$. Specifically,
  \[
    \widetilde A_Y^{N,\Delta t,P}-A^\infty
    =D_Y^{N,\Delta t,P}
    +\bigl(A_Y^{N,\Delta t}-A^\infty\bigr).
  \]
  The second term is $\mathcal F_Y$-measurable, whereas
  $\E[D_Y^{N,\Delta t,P}\mid\mathcal F_Y]=0$. Conditional orthogonality gives
  \[
    \E\|\widetilde A_Y^{N,\Delta t,P}-A^\infty\|_F^2
    =\E\|D_Y^{N,\Delta t,P}\|_F^2
    +\E\|A_Y^{N,\Delta t}-A^\infty\|_F^2.
  \]
  Bounding the first term by \eqref{eq:random-pair-design-moment} and the second
  by part~(ii) gives $\mathcal D^Y_{N,\Delta t,P}$. The operator-norm estimate
  again follows from $\|B\|_{\mathrm{op}}\le\|B\|_F$.
\end{proof}

The $1/N$ term in $\mathcal D_{N,\Delta t}$ includes fluctuations from the
random initial data and can remain when $\sigma=0$. A fixed positive
observation-noise level likewise leaves a nonvanishing term, so increasing $N$
alone does not establish convergence of $A_Y^{N,\Delta t}$ to $A^\infty$.

\subsection{From Continuum Identifiability to Coefficient Recovery}
\label{app:continuum}

The mean-square design estimates give high-probability bounds on the
deviation from $A^\infty$ by Markov's inequality. Subtracting the resulting
deviation radii from the continuum stability constants gives finite-design
gaps. We combine these gaps with the residual moments and deterministic
regression bounds to obtain coefficient recovery.

For $\delta_A\in(0,1)$ and $\gamma_\infty>0$, set
\begin{align}
  r^Y_{N,\Delta t,\delta_A}
   & :=\sqrt{\frac{\mathcal D^Y_{N,\Delta t}}{\delta_A}},
  \qquad g^Y_{N,\Delta t,\delta_A}:=\gamma_\infty-r^Y_{N,\Delta t,\delta_A},
  \label{eq:def-r-NdtdA} \\
  r^Y_{N,\Delta t,P,\delta_A}
   & :=\sqrt{\frac{\mathcal D^Y_{N,\Delta t,P}}{\delta_A}},
  \qquad g^Y_{N,\Delta t,P,\delta_A}:=\gamma_\infty-r^Y_{N,\Delta t,P,\delta_A}.
  \label{eq:def-random-pair-continuum-gap}
\end{align}
For $\gamma_{\infty,\mathrm{sp}}>0$ and coefficient changes in
$\mathcal C_S$, set
\begin{equation}
  g^{Y,\mathrm{sp}}_{N,\Delta t,\delta_A}
  :=\gamma_{\infty,\mathrm{sp}}-r^Y_{N,\Delta t,\delta_A},
  \qquad g^{Y,\mathrm{sp}}_{N,\Delta t,P,\delta_A}
  :=\gamma_{\infty,\mathrm{sp}}-r^Y_{N,\Delta t,P,\delta_A}.
  \label{eq:def-sparse-continuum-gaps}
\end{equation}
For the sampled oracle OLS bound, define the event
\begin{equation}
  \mathcal E^Y_{\infty,P,\delta_A}
  :=\{\sigma_{\min}(\widetilde A_{Y,S}^{N,\Delta t,P})
  \ge g^Y_{N,\Delta t,P,\delta_A}\}.
  \label{eq:def-random-pair-continuum-event}
\end{equation}

\begin{lemma}[Continuum singular-value transfer]
  \label{lem:continuum-singular-transfer}
  Suppose the assumptions of Proposition~\ref{prop:weak-design-convergence} and
  Assumption~\ref{ass:observation-noise} hold, with $r\ge3$ in
  Assumption~\ref{ass:test_functions_A}, and suppose that, for some
  $\gamma_\infty>0$,
  \[
    \sigma_{\min}(A_S^\infty)\ge\gamma_\infty.
  \]
  The all-pair and sampled gaps are
  defined in \eqref{eq:def-r-NdtdA} and
  \eqref{eq:def-random-pair-continuum-gap}, respectively. For every
  $\delta_A\in(0,1)$:
  \begin{enumerate}[label=(\roman*),leftmargin=2.2em]
    \item with probability at least $1-\delta_A$,
          \[
            \sigma_{\min}(A_{Y,S}^{N,\Delta t})
            \ge g^Y_{N,\Delta t,\delta_A};
          \]
    \item if Assumption~\ref{ass:random-pair-sampling} also holds, then with
          probability at least $1-\delta_A$,
          \[
            \sigma_{\min}(\widetilde A_{Y,S}^{N,\Delta t,P})
            \ge g^Y_{N,\Delta t,P,\delta_A}.
          \]
  \end{enumerate}
\end{lemma}

\begin{proof}[Proof of Lemma~\ref{lem:continuum-singular-transfer}]
  Proposition~\ref{prop:weak-design-convergence}(ii) and Markov's inequality give
  $\|A_Y^{N,\Delta t}-A^\infty\|_F\le r^Y_{N,\Delta t,\delta_A}$
  with probability at least $1-\delta_A$. Restricting to the active columns
  does not increase the operator norm, so
  \[
    \|A_{Y,S}^{N,\Delta t}-A_S^\infty\|_{\mathrm{op}}
    \le\|A_Y^{N,\Delta t}-A^\infty\|_{\mathrm{op}}
    \le\|A_Y^{N,\Delta t}-A^\infty\|_F.
  \]
  The standard singular-value perturbation inequality
  \[
    |\sigma_{\min}(A)-\sigma_{\min}(B)|
    \le\|A-B\|_{\mathrm{op}}
  \]
  together with $\|A-B\|_{\mathrm{op}}\le\|A-B\|_F$ gives part~(i).
  Part~(ii) follows identically from
  Proposition~\ref{prop:weak-design-convergence}(iii).
\end{proof}

\begin{lemma}[Continuum sparse-cone and column-norm transfer]
  \label{lem:continuum-sparse-transfer}
  Suppose the assumptions of Proposition~\ref{prop:weak-design-convergence} and
  Assumption~\ref{ass:observation-noise} hold, with $r\ge3$ in
  Assumption~\ref{ass:test_functions_A}, and suppose that, for some
  $\gamma_{\infty,\mathrm{sp}}>0$,
  \[
    \gamma_{\mathrm{sp}}(A^\infty;S)
    \ge\gamma_{\infty,\mathrm{sp}}.
  \]
  The sparse gaps are defined in
  \eqref{eq:def-sparse-continuum-gaps}. For every $\delta_A\in(0,1)$, with
  probability at least $1-\delta_A$,
  \[
    \gamma_{\mathrm{sp}}(A_Y^{N,\Delta t};S)
    \ge g^{Y,\mathrm{sp}}_{N,\Delta t,\delta_A},
    \qquad
    R(A_Y^{N,\Delta t})\le R_\infty+r^Y_{N,\Delta t,\delta_A}.
  \]
  If Assumption~\ref{ass:random-pair-sampling} also holds, then with
  probability at least $1-\delta_A$,
  \[
    \gamma_{\mathrm{sp}}(\widetilde A_Y^{N,\Delta t,P};S)
    \ge g^{Y,\mathrm{sp}}_{N,\Delta t,P,\delta_A},
    \qquad R(\widetilde A_Y^{N,\Delta t,P})
    \le R_\infty+r^Y_{N,\Delta t,P,\delta_A}.
  \]
\end{lemma}

\begin{proof}[Proof of Lemma~\ref{lem:continuum-sparse-transfer}]
  On the event
  $\|A_Y^{N,\Delta t}-A^\infty\|_F\le r^Y_{N,\Delta t,\delta_A}$,
  the reverse triangle inequality gives, for every $u\in\mathcal C_S$,
  \[
    \|A_Y^{N,\Delta t}u\|_2
    \ge\bigl(\gamma_{\infty,\mathrm{sp}}
    -r^Y_{N,\Delta t,\delta_A}\bigr)\|u\|_2.
  \]
  The same bound on $\|A_Y^{N,\Delta t}-A^\infty\|_F$ also controls
  $R(A_Y^{N,\Delta t})$. The event has
  probability at least $1-\delta_A$ by
  Proposition~\ref{prop:weak-design-convergence}(ii) and Markov's inequality.
  The conclusion for $\widetilde A_Y^{N,\Delta t,P}$ follows from
  Proposition~\ref{prop:weak-design-convergence}(iii) in the same way. In both cases $\|B\|_{\mathrm{op}}\le\|B\|_F$ is used.
\end{proof}

\begin{proposition}[Coefficient recovery after pair sampling]
  \label{prop:explicit-gap-random-pair-recovery}
  Suppose Assumptions~\ref{ass:particle-system},
  \ref{ass:observation-noise}, \ref{ass:dictionary-model},
  \ref{ass:test_functions_A}, \ref{ass:time_grid_A}, and
  \ref{ass:random-pair-sampling} hold, with $r\ge3$ in
  Assumption~\ref{ass:test_functions_A}, and assume in addition that
  $\E\|X_1(0)\|^4<\infty$. Let $\delta_A,\varepsilon\in(0,1)$ satisfy
  $\delta_A+\varepsilon<1$.
  \begin{enumerate}[label=(\roman*),leftmargin=2.2em]
    \item \textbf{Oracle OLS.}
          Suppose that, for some $\gamma_\infty>0$,
          \[
            \sigma_{\min}(A_S^\infty)\ge\gamma_\infty,
          \]
          and $g^Y_{N,\Delta t,P,\delta_A}>0$. Then, with probability at least
          $1-\delta_A-\varepsilon$,
          \[
            \|\hat\theta_{S,Y,P}-\theta_S^\star\|_2^2
            \le\frac{\mathcal V^Y_{N,\Delta t,P}}
            {\varepsilon\bigl(g^Y_{N,\Delta t,P,\delta_A}\bigr)^2}.
          \]
          Moreover, for the event in
          \eqref{eq:def-random-pair-continuum-event},
          \[
            \E\left[
              \|\hat\theta_{S,Y,P}-\theta_S^\star\|_2^2
              \mathbf1_{\mathcal E^Y_{\infty,P,\delta_A}}\right]
            \le\frac{\mathcal V^Y_{N,\Delta t,P}}
            {\bigl(g^Y_{N,\Delta t,P,\delta_A}\bigr)^2}.
          \]
    \item \textbf{LASSO.}
          Suppose that, for some $\gamma_{\infty,\mathrm{sp}}>0$,
          \[
            \gamma_{\mathrm{sp}}(A^\infty;S)
            \ge\gamma_{\infty,\mathrm{sp}},
          \]
          and $g^{Y,\mathrm{sp}}_{N,\Delta t,P,\delta_A}>0$. For any
          \[
            \lambda\ge2\bigl(R_\infty+r^Y_{N,\Delta t,P,\delta_A}\bigr)
            \sqrt{\frac{\mathcal V^Y_{N,\Delta t,P}}{\varepsilon}},
          \]
          let $\hat\theta_{Y,P}^\lambda$ be any minimizer of
          \eqref{eq:def-random-pair-LASSO}. Then, with probability at least
          $1-\delta_A-\varepsilon$,
          \[
            \|(\hat\theta_{Y,P}^\lambda-\theta^\star)_{S^c}\|_1
            \le3\|(\hat\theta_{Y,P}^\lambda-\theta^\star)_S\|_1,
          \]
          and
          \[
            \|\hat\theta_{Y,P}^\lambda-\theta^\star\|_2^2
            \le\frac{9s\lambda^2}
            {(g^{Y,\mathrm{sp}}_{N,\Delta t,P,\delta_A})^4}.
          \]
          With equality in the lower bound on $\lambda$ above,
          \[
            \|\hat\theta_{Y,P}^\lambda-\theta^\star\|_2^2
            \le
            \frac{36s(R_\infty+r^Y_{N,\Delta t,P,\delta_A})^2}
            {\varepsilon(g^{Y,\mathrm{sp}}_{N,\Delta t,P,\delta_A})^4}
            \mathcal V^Y_{N,\Delta t,P}.
          \]
  \end{enumerate}
\end{proposition}

\begin{proof}[Proof of Proposition~\ref{prop:explicit-gap-random-pair-recovery}]
  Proposition~\ref{prop:residual-design-estimates}(ii) supplies the residual
  second-moment bound $\mathcal V^Y_{N,\Delta t,P}$. For part~(i), this bound
  and the design event from Lemma~\ref{lem:continuum-singular-transfer}(ii)
  verify the hypotheses of Proposition~\ref{prop:generic-oracle}, which gives
  both OLS conclusions.

  For part~(ii), Markov's inequality applied to the same residual bound gives
  $\|\widetilde\eta_Y^{N,\Delta t,P}\|_2
  \le\sqrt{\mathcal V^Y_{N,\Delta t,P}/\varepsilon}$ with probability at least
  $1-\varepsilon$. Intersect this event with the design event from
  Lemma~\ref{lem:continuum-sparse-transfer}. On their intersection,
  Cauchy--Schwarz and the column-norm bound give
  \[
    2\|(\widetilde A_Y^{N,\Delta t,P})^\top
    \widetilde\eta_Y^{N,\Delta t,P}\|_\infty
    \le2\bigl(R_\infty+r^Y_{N,\Delta t,P,\delta_A}\bigr)
    \sqrt{\frac{\mathcal V^Y_{N,\Delta t,P}}{\varepsilon}}
    \le\lambda.
  \]
  Thus the score event in \eqref{eq:def-lasso-score-event} and sparse-cone
  stability hold simultaneously.
  Proposition~\ref{prop:generic-lasso} gives the LASSO conclusions, and a union
  bound gives probability at least
  $1-\delta_A-\varepsilon$.
\end{proof}

\begin{proof}[Proof of Theorem~\ref{thm:random-pair-recovery}]
  For oracle OLS, the sampled-gap definition
  \eqref{eq:def-random-pair-continuum-gap} and the displayed design-error
  condition in Theorem~\ref{thm:random-pair-recovery}(i) imply
  \[
    r^Y_{N,\Delta t,P,\delta_A}\le\frac{\gamma_\infty}{2},
    \qquad
    g^Y_{N,\Delta t,P,\delta_A}\ge\frac{\gamma_\infty}{2}.
  \]
  Substitution into
  Proposition~\ref{prop:explicit-gap-random-pair-recovery}(i) proves the
  oracle OLS bound~\eqref{eq:main-oracle-recovery-bound}. For LASSO, the design-error condition in
  Theorem~\ref{thm:random-pair-recovery}(ii) gives
  \[
    g^{Y,\mathrm{sp}}_{N,\Delta t,P,\delta_A}
    \ge\frac{\gamma_{\infty,\mathrm{sp}}}{2},
    \qquad
    R_\infty+r^Y_{N,\Delta t,P,\delta_A}
    \le R_\infty+\frac{\gamma_{\infty,\mathrm{sp}}}{2}.
  \]
  By the sparse-gap definition \eqref{eq:def-sparse-continuum-gaps}, the
  penalty condition in Theorem~\ref{thm:random-pair-recovery}(ii) therefore
  implies the penalty condition in
  Proposition~\ref{prop:explicit-gap-random-pair-recovery}(ii). Substituting
  $g^{Y,\mathrm{sp}}_{N,\Delta t,P,\delta_A}
    \ge\gamma_{\infty,\mathrm{sp}}/2$ into that estimate proves the
  two inequalities in~\eqref{eq:main-lasso-recovery-bounds}. In particular, choosing the smallest admissible
  penalty, that is, equality in~\eqref{eq:random-pair-main-penalty}, also gives
  \[
    \|\hat\theta_{Y,P}^\lambda-\theta^\star\|_2^2
    \le
    \frac{576s(R_\infty+\gamma_{\infty,\mathrm{sp}}/2)^2}
    {\varepsilon\gamma_{\infty,\mathrm{sp}}^4}
    \mathcal V^Y_{N,\Delta t,P}.
  \]
  This completes the proof.
\end{proof}

\section{Additional Numerical Experiments}
\label{app:additional-numerics}

This appendix specifies the numerical estimators, error metrics, and complete protocols for the one-dimensional, predator--prey, and ten-dimensional experiments.

\subsection{Numerical Setup and Evaluation}
\label{sec:numerical-setup}
\paragraph{Test functions.}
The experiments use the test counts and composite trapezoidal quadrature specified in Section~\ref{sec:numerical-experiments}. The isotropic Gaussian spatial tests are
\[
  \varphi_\ell(x)
  =\exp\left(-\frac{\|x-c_\ell\|^2}{2s_x^2}\right),
  \qquad
  c_\ell\in\R^d,\quad s_x>0,
\]
where $c_\ell$ is the spatial center of the $\ell$th test and $s_x$ is the common width. In the 10D demonstration, we draw the centers uniformly from the coordinatewise box determined by the empirical $1\%$ and $99\%$ quantiles of the observed positions.

The smooth temporal bumps have unit peak:
\[
  \psi_k(t)=\exp\left(1-\frac{1}{1-4(t-\tau_k)^2/H_t^2}\right)
\]
for $|t-\tau_k|<H_t/2$, and we set $\psi_k(t)=0$ otherwise. Here
$\tau_k=kT/(L_t+1)$ for $k=1,\ldots,L_t$ gives the uniformly spaced centers,
and $H_t$ is the temporal-window length.

\paragraph{Regression variants.}
Weak regression uses the time-integrated test equations in Section~\ref{sec:regression-representation}; strong FD regression instead fits particlewise forward position differences divided by the observation interval. For either design, OLS is unregularized minimum-norm least squares, Ridge adds an $\ell_2$ penalty, and LASSO adds an $\ell_1$ penalty \citep{Tibshirani1996Regression}. All-pair weak fits replace $\widetilde A_Y^{N,\Delta t,P}$ by $A_Y^{N,\Delta t}$ in~\eqref{eq:def-random-pair-LASSO}.

\paragraph{Error metrics and radial convention.}
We evaluate coefficient recovery, the induced particle drift, and coupled
trajectories at the penalty fixed for each experiment. A scalar profile defines
a vector atom by
\begin{equation}
  \Phi_m(z)=z\,\phi_m(\|z\|).
  \label{eq:numerical-radial-convention}
\end{equation}
For experiments with a radial true kernel, we write
$K^\star(z)=z\,\phi(\|z\|)$. The recovered profile is
$\hat\phi(r):=\sum_{m=1}^M\hat\theta_m\phi_m(r)$, so
$K_{\hat\theta}(z)=z\,\hat\phi(\|z\|)$. Figures showing the interaction against
pairwise distance $r$ plot $r\phi(r)$ and $r\hat\phi(r)$.

For every realizable experiment, the coefficient error is
\[
  \frac{\|\hat\theta-\theta^\star\|_2}{\|\theta^\star\|_2}.
\]
For each target species $\alpha$, let $v_\alpha^\star$ and $\hat v_\alpha$ stack
all true and learned drifts evaluated on the held-out target--time points. If
$n_{\mathrm{tar}}$ is the number of target species, the drift error is
\begin{equation}
  \left\{\frac1{n_{\mathrm{tar}}}\sum_{\alpha=1}^{n_{\mathrm{tar}}}
  \frac{\|\hat v_\alpha-v_\alpha^\star\|_2^2}
       {\|v_\alpha^\star\|_2^2}\right\}^{1/2}.
  \label{eq:numerical-drift-error}
\end{equation}
For a single species, $n_{\mathrm{tar}}=1$. For the predator--prey system, the
average gives equal weight to prey and predator rather than allowing the $N$
prey velocities to dominate the single predator velocity. Both coefficient
and drift errors are dimensionless and relative. Across repeated runs, we
report their root-mean-square values.

For trajectory reproduction, let $X=(X_i(t_n))_{i,n}$ denote the latent
position array. We obtain $\hat X$ by simulating \eqref{eq:SDE} with the fitted
kernel in place of $K^\star$, using the same initial positions, Brownian
increments, and time grid. The trajectory error is
$\|\hat X-X\|_F/\|X\|_F$ for a single species; the predator--prey experiment
uses the analogous equal weighting of the two target species.

To isolate pair sampling, $\hat\theta$ and $\hat\theta_P$ denote estimates
computed with $A_Y^{N,\Delta t}$ and $\widetilde A_Y^{N,\Delta t,P}$,
respectively. Figure~\ref{fig:validation-sweeps}(e) reports the root-mean-square
of $\|\hat\theta_P-\hat\theta\|_2/\|\theta^\star\|_2$ across runs.

\paragraph{Regression implementation.}
The weak systems use the $1/\sqrt{L_xL_t}$ normalization in \eqref{eq:def-bY-dt}--\eqref{eq:def-AY-dt}. We use raw columns and responses throughout, without centering, standardization, or an intercept. Minimum-norm OLS is computed with \texttt{numpy.linalg.lstsq} using \texttt{rcond=None}; some six-atom zero-penalty systems use the equivalent \texttt{sklearn.linear\_model.Ridge} implementation. Ridge uses an SVD solver and a candidate set containing zero and 60 positive penalties from $10^{-10}\|A\|_{\mathrm{op}}^2$ to $10^2\|A\|_{\mathrm{op}}^2$. LASSO is computed with \texttt{sklearn.linear\_model.Lasso}, no intercept, and cyclic coordinate descent. For weak systems, the software penalty satisfies $\lambda=L_xL_t\alpha_{\mathrm{sk}}$ in \eqref{eq:def-random-pair-LASSO}.

Independent validation trajectories select penalties by minimizing the residual loss for the corresponding weak or strong FD design. The experiment-specific LASSO grids are given below; Ridge uses the grid above. The predator--prey prey and predator equations select their penalties separately. Penalty selection uses only validation residuals; it does not use true coefficients, coefficient errors, or drift and trajectory errors evaluated on held-out data. In the 10D pair-count sweep, validation at the nominal $P_0=25000$ selects one absolute weak LASSO penalty that is then fixed across all pair counts. We report the selected solver coefficients directly and aggregate every repetition.

\paragraph{Comparison across the three experiments.}
\label{app:cross-example-comparison}
Table~\ref{tab:cross-example-comparison} reports coefficient and drift RMSEs for six methods: weak LASSO, weak Ridge, weak OLS, strong FD--LASSO, strong FD--Ridge, and strong FD--OLS. Methods are evaluated on the same independently generated repetitions and remain paired within each repetition. All three experiments use independent Gaussian position noise with $\sigma_{\mathrm{obs}}=0.01$. Within each example, all six methods use the same observed training trajectories. The weak methods use $P=50N$ sampled pairs, whereas the strong FD methods fit particlewise forward differences using full source averages and do not use the weak pair design.

The 1D Newtonian-repulsion comparison has 100 repetitions. The 2D predator--prey comparison combines all three directed coefficient vectors over 20 repetitions. Its strong FD prey regression stacks particlewise forward differences from the four training trajectories in a 12-column system, while the predator regression uses a separate five-column system; penalized prey and predator fits use separately selected validation penalties. The 10D distance-localized comparison uses $N=500$, process noise $\sigma=0.05$, observation interval $0.02$, and 20 repetitions. Its spatial-test centers are determined from the corresponding observed training or validation positions. The strong FD design uses 64 target particles and the full 500-particle source average at every observation time. OLS uses a minimum-norm solution on the raw design columns, and all penalized fits use independent validation data.

For the Table~\ref{tab:cross-example-comparison} comparisons, weak LASSO uses 60 logarithmically spaced penalties from the design-specific $\alpha_{\max}$ to $10^{-7}\alpha_{\max}$, while strong FD--LASSO uses 60 penalties down to $10^{-9}\alpha_{\max}$. Weak Ridge and strong FD--Ridge use zero together with the 60 positive spectral-scale penalties specified above. Each penalty minimizes the residual loss of its own formulation on independent validation data; the predator--prey prey and predator systems are selected separately.

For drift evaluation, 1D uses 2000 target-particle/time combinations sampled once from one independent trajectory; 2D and 10D use an independent latent test trajectory in each repetition. Each evaluation is shared across methods and excluded from fitting and penalty selection.

\subsection{Additional Settings for the One-Dimensional Study}
\label{app:one-dimensional-setup}

For all 1D experiments in this subsection, we fix the Gaussian test centers in advance, equally spaced on $[0,4]$, and use $s_x=0.25$ and $H_t=1.8T/(L_t+1)$. At each observation time, one sampled pair set is shared across tests and atoms, while training and validation use independent pair streams. We draw the initial particle positions independently from an equally weighted mixture of three Gaussian distributions with means $0.5$, $2.0$, and $3.5$ and common standard deviation $0.005$.

\paragraph{1D Newtonian-repulsion experiment.}
We use the six-atom dictionary and coefficient vector specified in Section~\ref{sec:num-newtonian-repulsion-experiment}. All six vector atoms under~\eqref{eq:numerical-radial-convention} are globally Lipschitz. The fit is exactly realizable and sparse, and both estimators use the complete dictionary.

The observation-noise comparison uses $N=100$, $T=2$, $\Delta t=0.005$, $\sigma=0.1$, $P=50N$, and 100 independent Euler--Maruyama training trajectories. Observation-noise levels share each latent trajectory and a standardized Gaussian noise array, making their comparison paired. All methods share the 100 training trajectories and observed positions. Weak LASSO uses a separate validation trajectory and independent validation pair stream, while strong FD--LASSO selects its penalty on the corresponding forward-difference system from that independent validation trajectory.

The remaining six-atom experiments use exact positions, $\sigma_{\mathrm{obs}}=0$. The process-noise and particle-number sweeps use Euler--Maruyama with $T=2$, $\Delta t=0.001$, and $P=50N$. We use 100 independent training and validation trajectory pairs at every setting, varying $\sigma$ at fixed $N=100$ and varying $N$ at fixed $\sigma=0.1$. Figure~\ref{fig:validation-sweeps}(a)--(d) reports the coefficient RMSE across 100 paired repetitions; their bands are 95\% delta-method intervals obtained from the mean squared coefficient error. The observation-interval experiment sets $\sigma=\sigma_{\mathrm{obs}}=0$ to remove fixed noise terms from the comparison. We integrate the deterministic system by forward Euler with the simulation step equal to the observation interval, stack the weak equations from three independent training trajectories, and use a separate three-trajectory validation system with $N=200$, $T=10$, and $P=50N=10000$. The three training and three validation initial states in each repetition are paired across observation intervals.

For each sampled training system underlying Figure~\ref{fig:validation-sweeps}(a)--(d), we select the weak LASSO penalty from 60 logarithmically spaced values between $\alpha_{\max}$ and $10^{-7}\alpha_{\max}$ by minimizing the same weak squared loss on its independently sampled validation system. Cyclic coordinate descent uses at most 200000 iterations and tolerance $10^{-8}$. All 2200 selected fits converge.

The pair-count experiment uses one fixed Euler--Maruyama trajectory with $N=200$, $T=2$, $\Delta t=0.01$, and $\sigma=0.1$. For each repetition, an independent sampled validation system at $P_0=10000$ selects one absolute weak LASSO penalty, which is then fixed for every $P\in\{10^2,3\times10^2,10^3,3\times10^3,10^4,3\times10^4\}$ and also used for the all-pair reference. At each $P$, we perform 100 weak LASSO and weak OLS fits using $\widetilde A_Y^{N,\Delta t,P}$, compare each with its counterpart using $A_Y^{N,\Delta t}$, and resample the index pairs independently at every contributing time. Training and validation pair streams are independent, and all 600 fixed-penalty weak LASSO fits converge. The bands in Figure~\ref{fig:validation-sweeps}(e) are 95\% intervals over the 100 pair-sampling seeds. Finally, the estimator-independent timing comparison uses all-pair and $P=50N$ sampled construction over ten independent process invocations. The plotted center is the median and the bands are interquartile ranges; simulation, construction of $b_Y^{N,\Delta t}$, regression, diagnostics, and file I/O are excluded.

\paragraph{Generic piecewise-linear dictionary experiment.}
For the dictionary defined in Section~\ref{sec:num-hat-approximation}, we use $M\in\{9,11,14,17,22,27,33,42,52,65\}$. Its constant-extended vector atoms are globally Lipschitz. Because the true profile lies outside every finite span, a finite true coefficient vector and coefficient error are unavailable. We therefore compare dictionary sizes using the drift error in~\eqref{eq:numerical-drift-error}, evaluated on an independent latent test trajectory with all particles serving as sources and targets. It equals one for the zero-drift estimate.

These finite dictionaries belong to a standard approximation hierarchy. Let $V_h$ be the continuous piecewise-linear space on a uniform mesh of width $h$ in $[0,4]$. For every continuous radial profile $\phi$ and $K(z)=z\phi(|z|)$, there are $\phi_h\in V_h$ such that, as $h\to0$,
\begin{equation}
  \sup_{|z|\le4}|z|\,|\phi_h(|z|)-\phi(|z|)|
  \le 4\|\phi_h-\phi\|_\infty\longrightarrow0.
  \label{eq:hat-density}
\end{equation}
This compact-domain density statement gives the approximation limit on $[0,4]$; uniform kernel approximation also controls the induced drift on configurations in this interval. For $\varepsilon_K=10^{-8}$, the true profile has curvature $\phi''(r)=2/(r+\varepsilon_K)^3$, which is extremely large near zero and makes a whole-interval interpolation bound highly conservative.

For every $M$, each of the 100 repetitions uses separate Euler--Maruyama training, validation, and latent test trajectories with $N=1000$, $T=2$, $\Delta t=0.005$, $\sigma=0.1$, $\sigma_{\mathrm{obs}}=0.01$, and $P=50N=50000$. The seed triplets are shared across dictionary sizes, so the comparison is paired. Independent Gaussian observation noise is added to the training and validation trajectories, while drift and trajectory errors are evaluated on the latent test trajectory. Weak LASSO and strong FD--LASSO each use 60 logarithmically spaced penalties from the design-specific $\alpha_{\max}$ to $10^{-9}\alpha_{\max}$. We minimize the corresponding residual loss on the independent validation system among converged candidates and then evaluate the selected coefficients on the test trajectory. Cyclic coordinate descent uses at most $200000$ iterations and tolerance $10^{-8}$.

The visualizations in Figure~\ref{fig:one-dimensional-hat-practical}(a,b) use the displayed fit at $M=14$, the dictionary size with the smallest mean weak LASSO drift error. This choice determines only the interaction and trajectory visualizations. Figure~\ref{fig:one-dimensional-hat-practical}(c) reports every dictionary size: gray points show all weak LASSO drift errors, and the three curves show arithmetic means with percentile-bootstrap 95\% confidence intervals from 10000 resamples. All 3000 selected coefficient estimates and drift evaluations are finite. The weak OLS design is rank deficient in all 1000 fits. Weak LASSO has the smallest mean drift error at eight dictionary sizes, strong FD--LASSO at $M=9$, and weak OLS at $M=11$.

\subsection{Two-Dimensional Predator--Prey Experiment}
\label{sec:num-predator-prey}

We adapt the weak system to the deterministic predator--swarm model of \citet{ChenKolokolnikov2014Predator}. For prey positions $X_i\in\R^2$ and one predator position $Z\in\R^2$, we use the smooth regularization
\begin{align}
 \dot X_i
 &=\frac1N\sum_{j\ne i}\left[
 \frac{X_i-X_j}{\|X_i-X_j\|^2+\delta^2}-a(X_i-X_j)
 \right]
 +b\frac{X_i-Z}{\|X_i-Z\|^2+\delta^2},
 \label{eq:predator-prey-prey}\\
 \dot Z
 &=\frac cN\sum_{j=1}^N
 \frac{X_j-Z}{(\|X_j-Z\|^2+\delta^2)^{p_{\mathrm{pred}}/2}},
 \label{eq:predator-prey-predator}
\end{align}
with $a=1$, $b=0.2$, $c=0.4$, $p_{\mathrm{pred}}=3$, and $\delta=0.05$. These parameters produce the ring state of the original model while the positive $\delta$ makes every fitted atom smooth and globally Lipschitz.

The prey weak equation has separate blocks for prey--prey and predator-on-prey interactions. Each block uses the six scalar profiles
\[
 (r^2+\delta^2)^{-1},\quad 1,\quad
 e^{-r/0.25},\quad e^{-r/0.5},\quad e^{-r},\quad e^{-r/2}.
\]
Their true coefficient vectors are $(1,-1,0,0,0,0)^\top$ and $(0.2,0,0,0,0,0)^\top$. The predator equation uses
\[
 (r^2+\delta^2)^{-q/2},\qquad q\in\{1,2,3,4,5\},
\]
with true coefficients $(0,0,0.4,0,0)^\top$. For each target species, the corresponding weak design averages only over the source species before contraction with the target test gradient. For target species $\alpha$ and source species $\beta$, with sizes $N_\alpha$ and $N_\beta$, the corresponding design block has entries
\[
 (A_{\alpha\beta})_{a,m}=\frac{1}{\sqrt L\,N_\alpha N_\beta}
 \sum_{i\in \alpha}\sum_{j\in \beta}^{\prime}\sum_n w_n\psi_k(t_n)
 \nabla\varphi_\ell(Y_i^{\alpha,n})\cdot\Phi_m^{\alpha\beta}(\xi_{ij}^{\alpha\beta,n}),
\]
where $\xi_{ij}^{\alpha\beta,n}$ follows the directed displacement in
\eqref{eq:predator-prey-prey}--\eqref{eq:predator-prey-predator} (in
particular, $X_j^n-Z^n$ for $\alpha=Z,\beta=X$), and the prime excludes $j=i$ only
for a within-species block. The response
$b_\alpha$ uses the same $1/(\sqrt L N_\alpha)$ target average as in
\eqref{eq:def-bY-dt}. Writing $b_X$ and $b_Z$ for the stacked prey and predator weak responses, we obtain the following block form for the complete regression:
\begin{equation}
 \begin{bmatrix} b_X\\ b_Z \end{bmatrix}
 \simeq
 \begin{bmatrix}
  A_{XX} & A_{XZ} & 0\\
  0 & 0 & A_{ZX}
 \end{bmatrix}
 \begin{bmatrix} \theta_{XX}\\ \theta_{XZ}\\ \theta_{ZX} \end{bmatrix}.
 \label{eq:predator-prey-block-regression}
\end{equation}
Thus the prey response gives one 12-column row block for $(\theta_{XX},\theta_{XZ})$, while the predator response gives one five-column row block for $\theta_{ZX}$. We fit these two row blocks separately so that independent validation selects one penalty for each response; no channel coefficient is supplied to either fit.

Each of 20 repetitions has four training, four validation, and one test trajectory with $N=100$, $T=3$, observation interval $\Delta t=0.01$, $\sigma_{\mathrm{obs}}=0.01$, and $P=50N=5000$ for the prey--prey block. One pair set is shared across tests and atoms at each observation time, training and validation use independent pair streams, and both cross-species sums are exact. Fourth-order Runge--Kutta trajectories share a $0.005$ simulation grid and are subsampled at the observation interval. The weak nominal fits are exactly the $\sigma_{\mathrm{obs}}=0.01$ records from the paired sweep below; the strong FD fits use the same observed trajectory arrays. The 25 Gaussian centers form the fixed grid $\{-1.5,-0.75,0,0.75,1.5\}^2$, with widths $0.5$ and $0.7$ for the prey and predator equations. Weak LASSO and strong FD--LASSO select among 60 penalties separately for the prey and predator equations on independent validation systems, and every selected fit is finite.

\begin{figure}
  \centering
  \includegraphics[width=0.9\linewidth]{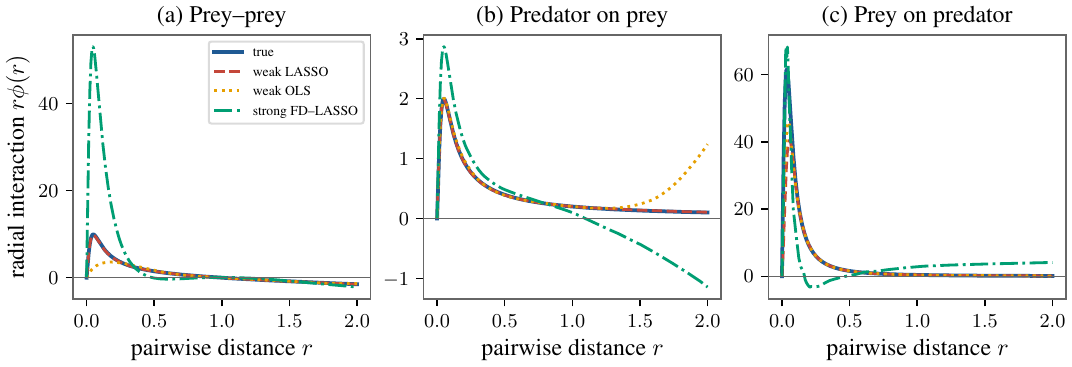}
  \caption{Median recovery of the three directed predator--prey interactions over 20 repetitions. Blue is the true interaction; red dashed, orange dotted, and green dash-dotted curves are weak LASSO, weak OLS, and strong FD--LASSO.}
  \label{fig:predator-prey-profiles}
\end{figure}

Figure~\ref{fig:predator-prey-profiles} shows close agreement between weak LASSO and the truth for the two prey-response channels and greater variation in the prey-on-predator channel, which is observed through one predator per trajectory. For weak LASSO and weak OLS, respectively, the median coefficient errors are $0.0976$ and $182$ for the joint prey vector, and $0.120$ and $1.74$ for the predator vector. The large weak OLS coefficient error reflects cancellation among correlated atoms. The corresponding median drift errors are $0.0603$ and $0.174$. Strong FD--LASSO has larger profile deviations than weak LASSO in all three channels.

\begin{figure}
  \centering
  \includegraphics[width=0.9\linewidth]{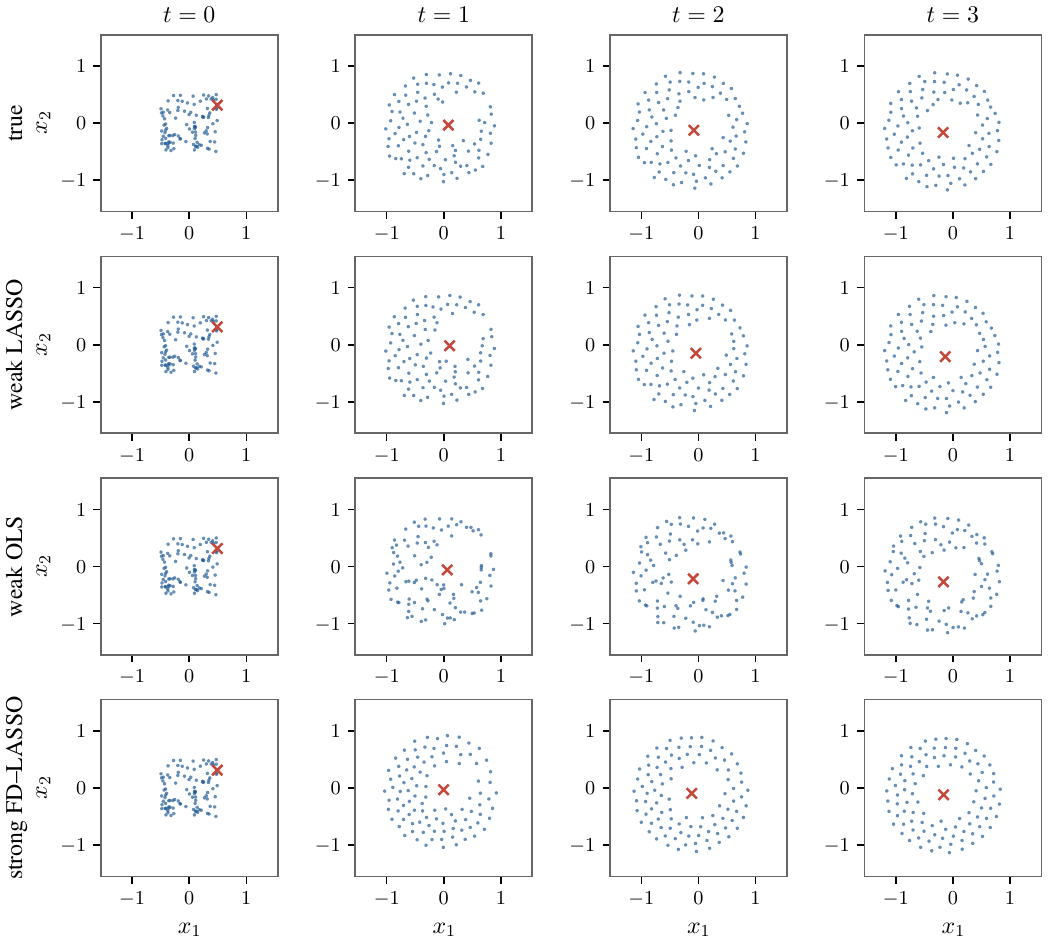}
  \caption{Representative predator--prey dynamics with the same 100-prey population used for fitting. Rows show the true, weak LASSO, weak OLS, and strong FD--LASSO systems from a shared initial condition over $[0,3]$. Blue points are prey and the red cross is the predator.}
  \label{fig:predator-prey-dynamics}
\end{figure}

Figure~\ref{fig:predator-prey-dynamics} shows that all three learned systems form an annular terminal swarm and track the predator over the observed horizon $[0,3]$. Their trajectory errors are $0.148$, $0.253$, and $0.216$ for weak LASSO, weak OLS, and strong FD--LASSO, respectively; weak LASSO gives the closest coupled trajectory.

\begin{figure}
  \centering
  \includegraphics[width=0.9\linewidth]{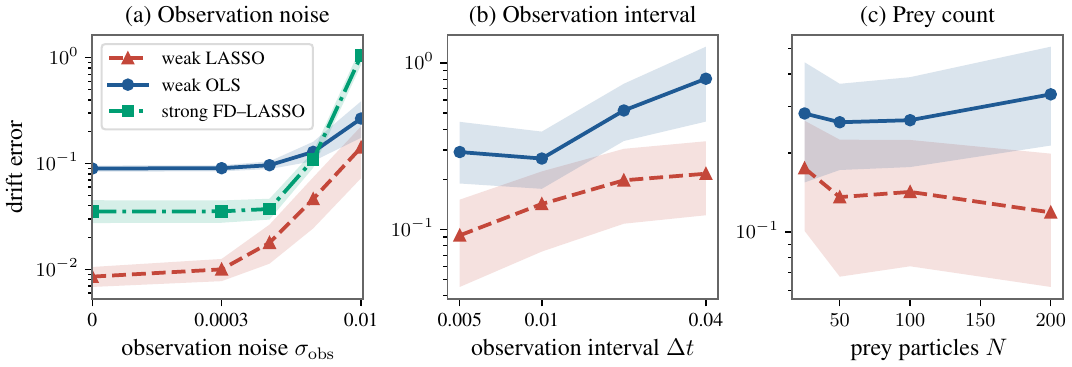}
  \caption{Drift error over 20 paired predator--prey repetitions versus (a) observation noise, (b) observation interval, and (c) prey count. Panel~(a) compares all three methods on the same trajectories; panels~(b--c) compare the weak methods at fixed $\sigma_{\mathrm{obs}}=0.01$. Lines show means, and bands give paired-bootstrap 95\% intervals.}
  \label{fig:predator-prey-robustness}
\end{figure}

We additionally repeat the complete fit 20 times at each observation-noise level $\sigma_{\mathrm{obs}}\in\{0,3\times10^{-4},10^{-3},3\times10^{-3},10^{-2}\}$, observation interval $\Delta t\in\{0.005,0.01,0.02,0.04\}$, and prey count $N\in\{25,50,100,200\}$. Seed indices are paired across conditions; the noise and observation-interval sweeps share latent trajectories, and the observation arrays are subsamples of simulations at step $0.005$. Training, validation, and latent test trajectories remain separate. We report the drift error in \eqref{eq:numerical-drift-error}. Weak LASSO and minimum-norm weak OLS use the same observed weak system, and Figure~\ref{fig:predator-prey-robustness}(a) applies strong FD--LASSO to the same observed arrays with an independently selected validation penalty. Curves show arithmetic means and 95\% paired percentile-bootstrap intervals from 10000 resamples.

In Figure~\ref{fig:predator-prey-robustness}(a), the log--log slopes over positive observation-noise levels are $0.766$, $0.304$, and $0.966$ for weak LASSO, weak OLS, and strong FD--LASSO. Strong FD--LASSO has mean drift error $0.0353$ for exact observations and $1.04$ at $\sigma_{\mathrm{obs}}=0.01$, while weak LASSO has the smaller mean throughout. In Figure~\ref{fig:predator-prey-robustness}(b), the observation-interval slopes are $0.416$ for weak LASSO and $0.535$ for weak OLS. At the common noise level in Figure~\ref{fig:predator-prey-robustness}(c), the weak LASSO mean changes from $0.175$ at $N=25$ to $0.119$ at $N=200$ but is not monotone between them, as the single-predator component limits scaling.

\subsection{Ten-Dimensional Distance-Localized Experiment}
\label{sec:num-localized-d10}

We use an experiment whose interaction is represented exactly by the dictionary and concentrated near two pairwise distances, separating high-dimensional estimation from dictionary misspecification. For centers
$\mu=(1.2,1.8,2.4,3.0)$ and width $h_{\mathrm{loc}}=0.2$, define
\begin{equation}
  K^\star(z)=z\sum_{m=1}^4\theta_m^\star
  \exp\!\left[-\frac{(\|z\|-\mu_m)^2}{2h_{\mathrm{loc}}^2}\right],
  \qquad \theta^\star=(0,1.5,-1,0)^\top.
  \label{eq:localized-d10-kernel}
\end{equation}
All estimators fit the complete four-profile distance-localized dictionary. We simulate $N=500$ particles in $\R^{10}$ for $500$ Euler--Maruyama steps with $\Delta t=0.01$ and baseline process noise $\sigma=0.05$, recording positions every two steps. Initial positions are i.i.d. from an equal-weight mixture of centered isotropic Gaussians with standard deviations $0.25$, $0.4$, $0.55$, and $0.7$, which supplies distances across the localized interaction regions.

Using the data-dependent spatial centers described above, we take $P=50N=25000$ sampled pairs. One pair set is shared across tests and atoms at each observation time, while training and validation use independent pair streams. Every recorded coordinate is corrupted independently by $\mathcal N(0,0.01^2)$ noise, and the spatial-test locations are recomputed from the corresponding observed training or validation trajectory. The radial interaction is displayed at 400 equally spaced radii on $[0.5,3.8]$. Drift errors use 64 targets at each of 12 equally spaced times on an independent latent trajectory, with all particles serving as sources. Weak LASSO selects among 60 penalties by minimizing the weak squared loss on an independent noisy validation trajectory, and minimum-norm weak OLS fits the identical four-column training system. Strong FD--LASSO uses the same observed trajectories, full source averages, and an independently selected forward-difference validation penalty. The simulation and fitting configuration was chosen in a separate three-repetition pilot; the results below use 20 new repetitions at the same $\sigma_{\mathrm{obs}}=0.01$ condition as the 10D row of Table~\ref{tab:cross-example-comparison}.

\begin{figure}
  \centering
  \includegraphics[width=0.9\linewidth]{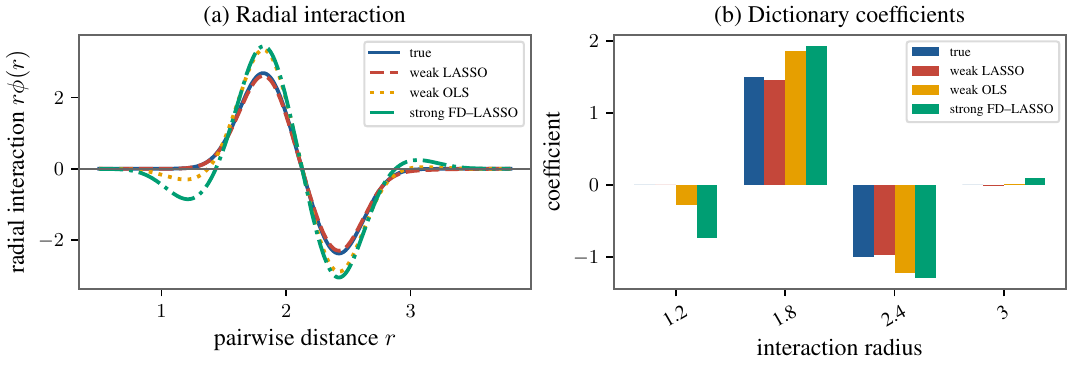}
  \caption{Representative ten-dimensional fit at $\sigma_{\mathrm{obs}}=0.01$: (a) true and estimated radial interactions $r\phi(r)$ against pairwise distance, and (b) their four coefficients, indexed by the interaction radii.}
  \label{fig:localized-d10-recovery}
\end{figure}

\begin{figure}
  \centering
  \includegraphics[width=0.9\linewidth]{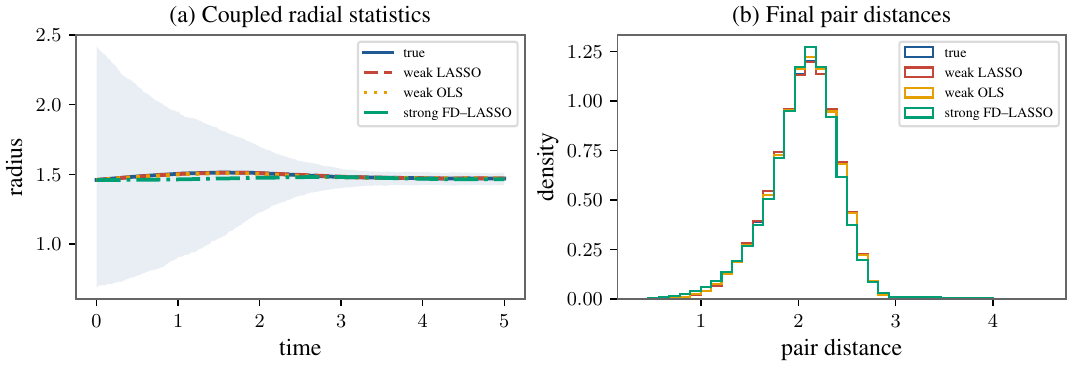}
  \caption{Coupled dynamics for the fits in Figure~\ref{fig:localized-d10-recovery}: (a) mean particle radius and the true 10--90\% particle band and (b) final pair-distance distributions. All systems use the same initial condition and Brownian increments.}
  \label{fig:localized-d10-dynamics}
\end{figure}

For the displayed repetition in Figure~\ref{fig:localized-d10-recovery}, weak LASSO has coefficient and drift errors $0.0310$ and $0.0274$. Its trajectory error through $T=5$ is $0.00479$ on an independent 500-particle rollout, matching the population size used for fitting. The corresponding weak OLS coefficient, drift, and trajectory errors are $0.280$, $0.188$, and $0.0283$, while the strong FD--LASSO errors are $0.498$, $0.255$, and $0.160$. Figure~\ref{fig:localized-d10-dynamics} shows the resulting differences in radial statistics and final pair-distance distributions.

We vary $\sigma\in\{0,0.05,0.075,0.1,0.15\}$ and $N\in\{100,200,350,500\}$. The pair counts are $P\in\{3,5,10,20,25,30\}\times10^3$. Seed indices are paired across conditions, while training, validation, and latent test trajectories remain separate. For the pair-count sweep, validation at the nominal $P_0=25000$ selects one absolute weak LASSO penalty per repetition, which is then fixed across all six training systems. At each observation time, the draws for smaller $P$ are prefixes of the common sampled stream, while draws at different observation times are independent.

\begin{figure}[tb]
  \centering
  \includegraphics[width=0.9\linewidth]{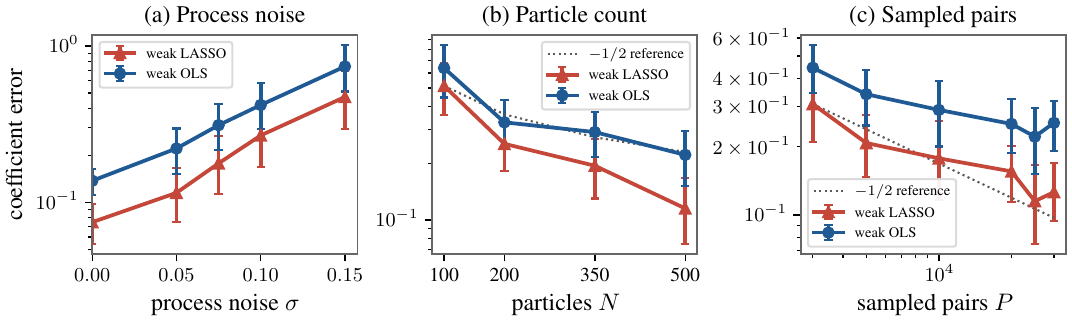}
  \caption{10D coefficient error versus (a) process noise, (b) particle count, and (c) sampled pairs at $\sigma_{\mathrm{obs}}=0.01$. Lines show means with paired-bootstrap 95\% intervals; dotted references have slope $-1/2$. Panel~(c) fixes the weak LASSO penalty.}
  \label{fig:localized-d10-robustness}
\end{figure}

At fixed observation noise, Figure~\ref{fig:localized-d10-robustness} shows the mean coefficient error for weak LASSO rising from $0.0747$ at $\sigma=0$ to $0.472$ at $\sigma=0.15$, with log--log slope $1.294$ over positive process-noise levels. It falls from $0.513$ at $N=100$ to $0.115$ at $N=500$, with slope $-0.873$, and from $0.308$ at $P=3000$ to $0.126$ at $P=30000$, with slope $-0.370$. Weak LASSO has lower mean coefficient error than weak OLS at every displayed condition.

\end{document}